\documentclass{article}

\usepackage[final]{neurips_2021}
\usepackage[utf8]{inputenc} 
\usepackage[T1]{fontenc}    

\usepackage{zshi}
\usepackage{microtype}

\title{Fast Certified Robust Training with Short Warmup}
\author{\normalsize
Zhouxing Shi\textsuperscript{1}\textsuperscript{*},
\enskip Yihan Wang\textsuperscript{1}\textsuperscript{*},
\enskip Huan Zhang\textsuperscript{1,2},
\enskip Jinfeng Yi\textsuperscript{3},
\enskip Cho-Jui Hsieh\textsuperscript{1}\\
{\normalsize \textsuperscript{1}UCLA \enskip \textsuperscript{2}CMU \enskip \textsuperscript{3}JD AI Research}\\
{\tt\small zshi@cs.ucla.edu, yihanwang@cs.ucla.edu, huan@huan-zhang.com,}\\
{\tt\small yijinfeng@jd.com, chohsieh@cs.ucla.edu}\\ 
{\small * Equal contribution}}
\date{}

\hypersetup{draft}
\begin{document}
\maketitle
\begin{abstract}
Recently, bound propagation based certified robust training methods have been proposed for training neural networks with certifiable robustness guarantees. Despite that state-of-the-art (SOTA) methods including interval bound propagation (IBP) and CROWN-IBP have per-batch training complexity similar to standard neural network training, they usually use a long warmup schedule with hundreds or thousands epochs to reach SOTA performance and are thus still costly. In this paper, we identify two important issues in existing methods, namely exploded bounds at initialization, and the imbalance in ReLU activation states and improve IBP training. These two issues make certified training difficult and unstable, and thereby long warmup schedules were needed in prior works. To mitigate these issues and conduct faster certified training with shorter warmup, we propose three improvements based on IBP training: 1) We derive a new weight initialization method for IBP training; 2) We propose to fully add Batch Normalization (BN) to each layer in the model, since we find BN can reduce the imbalance in ReLU activation states; 3) We also design regularization to explicitly tighten certified bounds and balance ReLU activation states during wamrup. We are able to obtain \textbf{65.03\%} verified error on CIFAR-10 ($\epsilon=\frac{8}{255}$) and \textbf{82.36\%} verified error on TinyImageNet ($\epsilon=\frac{1}{255}$) using very short training schedules (\textbf{160 and 80 total epochs}, respectively), outperforming literature SOTA trained with hundreds or thousands epochs under the same network architecture. The code is available at \url{https://github.com/shizhouxing/Fast-Certified-Robust-Training}.
\end{abstract}

\section{Introduction}

While deep neural networks (DNNs) are successfully applied in various areas, its robustness problem has attracted great attention since the discovery of adversarial examples \citep{szegedy2013intriguing,goodfellow2014explaining,carlini2017adversarial,kurakin2016adversarial,chen2017attacking,madry2017towards,su2018robustness,choi2019evaluating}, which poses concerns in DNN applications especially the safety-critical ones such as autonomous driving.
Methods for improving the empirical robustness of DNNs, such as adversarial training~\citep{madry2017towards}, provide no provable robustness guarantees, and thus some recent works aim to pursue \emph{certified robustness}. Specifically, the robustness is  evaluated in a certifiable manner using robustness verifiers~\citep{katz2017reluplex,zhang2018efficient,wong2018provable,singh2018fast,singh2019abstract, rudy2017, raghunathan2018semidefinite,wang2018efficient,xu2020automatic,wang2021beta}, which verify whether the model is provably robust against all possible input perturbations within the range. This is achieved usually by efficiently computing the output bounds.

To improve certified robustness, \emph{certified robust training} methods (also referred to as certified defense) minimize a certified robust loss computed by a verifier, and the certified loss is an upper bound of the worst-case loss given specified input perturbations. So far, Interval Bound Propagation (IBP)~\citep{gowal2018effectiveness,mirman2018differentiable} and CROWN-IBP~\citep{zhang2019towards,xu2020automatic} are the most efficient and effective methods for general models. IBP computes an interval with the output lower and upper bounds for each neuron, and CROWN-IBP further combines IBP with tighter linear relaxation-based bounds~\citep{zhang2018efficient,singh2019abstract} during warmup.

Both IBP and CROWN-IBP with loss fusion~\citep{xu2020automatic} have a per-batch training time complexity similar to standard DNN training. However, certified robust training remains costly and challenging, mainly due to their unstable training behavior -- they could easily diverge or stuck at a degenerate solution without a long ``warmup'' schedule.
The warmup schedule here refers to training the model with a regular (non-robust) loss first and then gradually increasing the perturbation radius from 0 to the target value in the robust loss (some previous works also refer to it as ``ramp-up'').
For example, generalized CROWN-IBP in \citet{xu2020automatic} used 900 epochs for warmup and 2,000 epochs in total to train a convolutional model on CIFAR-10~\citep{cifar}.

In this paper, we identify two important issues in existing certified training, so that a long warmup schedule could not be easily removed in previous works.
First, we find that the certified bounds can explode at the start of training, which is partly due to the suboptimal \emph{weight initialization} in prior works.
A good weight initialization is important for successful DNN training~\citep{Xavier,He_2015_ICCV}, but prior works for certified training generally use weight initialization methods originally designed for standard DNN training, while certified training is essentially optimizing a different type of augmented network defined by robustness verification~\citep{zhang2019towards}. 
The long warmup with gradually increasing perturbation radii in prior works can somewhat be viewed as finding a better initialization for final IBP training with the target radius, but it is too costly.
Second, we also observe that \emph{IBP leads to imbalanced ReLU activation states}, where the model prefers inactive (dead) ReLU neurons significantly more than other states because inactive neurons tend to tighten IBP bounds.  It can however hamper classification performance if too many neurons are dead. This issue can become more severe if the warmup schedule is shorter.

We focus on improving IBP training, since IBP is efficient per batch, and it is also the base of recent state-of-the-art methods~\citep{zhang2019towards,xu2020automatic}. We propose the following improvements:
\begin{itemize}[leftmargin=10pt]
	\item We derive a new weight initialization, \emph{IBP initialization}, for IBP-based certified training. The new initialization can stabilize the tightness of certified bounds at initialization.
	\item  We identify the benefit of Batch Normalization (BN) in certified training, and we find BN which normalizes pre-activation outputs can balance ReLU activation states and also stabilize variance. We propose to fully add BN to every layer, while it was partly or fully missed in prior works.
	\item We further propose regularizers to explicitly stabilize certified bounds and balance ReLU activation states during warmup. 
\end{itemize}

We are able to efficiently train certifiably robust models that outperform previous SOTA performance in significantly shorter training epochs. 
We achieve a verified error of \textbf{65.03\%} ($\epsilon=\frac{8}{255}$) on CIFAR-10 in \textbf{160} total training epochs, and \textbf{82.36\%} on TinyImageNet ($\epsilon=\frac{1}{255}$) in \textbf{80} epochs, based on efficient IBP training. 
Under the same convolution-based architecture, 
we significantly reduce the total training cost by $20\sim 60$ times compared to previous SOTA~\citep{zhang2019towards,xu2020automatic} or concurrent work~\citep{lyu2021evaluating}.
\section{Background and Related Work}
\label{sec:related_work}

\subsection{Certified Robust Training}
Training robust neural networks can generally be viewed as solving the following min-max optimization problem:
\begin{equation}
	\min_{\theta} \E_{(\rvx, y) \in \mathcal{X}} \left [ \max_{\delta\in\Delta(\rvx)} L(f_\theta(\rvx+\delta),y) \right ],
	\label{eq:minmax_opt}
\end{equation}
where $f_\theta$ stands for a neural network parameterized by $\theta$, $\mathcal{X}$ is the data distribution, $\rvx$ is a data example, $y$ is its ground-truth label, $\delta$ is a perturbation constrained by specification $\Delta(\rvx)$,   and $L$ is the loss function.
Empirical \emph{adversarial training} methods~\citep{goodfellow2014explaining,madry2017towards} solve the inner maximization in \eqref{eq:minmax_opt} with adversarial attack, and then solve the outer minimization as regular DNN training but augmented with $\delta$.
However, in adversarial training, the inner maximization has no guarantee to find a $\delta$ which can lead to worst model performance. 
In contrast, \emph{certified robust training} methods compute a certified upper bound for the inner maximization, so that the upper bound provably covers the worst-case perturbation.

In terms of certified robustness works,
\citet{raghunathan2018certified} used semidefinite relaxations for small two-layer models, and 
\citet{wong2018provable,mirman2018differentiable,dvijotham2018training,wang2018mixtrain} used linear relaxations but are still too computationally expensive for large models.
On the other hand, \citet{mirman2018differentiable} first used interval bounds to train a certifiably robust network, and \citet{gowal2018effectiveness} made it more effective. This approach is often referred to as interval bound propagation (IBP).
CROWN-IBP~\citep{zhang2019towards} further combined IBP with tighter linear relaxation bounds by CROWN~\citep{zhang2018efficient} during warmup, and it is generalized and accelerated in \citet{xu2020automatic}.
Additionally, \citet{balunovic2020adversarial} combined certified training with adversarial training; 
\citet{xiao2018training} added a ReLU stability regularizer to empirical adversarial training, to reduce unstable neurons for faster and tighter verification when tested with mixed integer programming (MIP), but their objective is distinct from ours and this method was shown not to improve certified training~\citep{lee2021loss}.
In concurrent works, \citet{lyu2021evaluating} proposed a parameterized ramp function as an alternative activation function, and used a tighter linear bound propagation algorithm for verification; \citet{zhang2021towards} proposed to use a different architecture with ``$\ell_{\infty}$-distance  neurons'' instead of traditional linear or convolutional layers. Yet they still need long training schedules. 

Moreover, while our scope in this paper is deterministic certified robustness, there are also randomization based works for probabilistic certified defense~\citep{cohen2019certified,li2018second,lecuyer2019certified,salman2019provably}. But randomized smoothing requires costly sampling at test time, and it is usually for $\ell_2$ perturbations and has fundamental limitations for $\ell_\infty$ ones~\citep{yang2020randomized,blum2020random,kumar2020curse}.

\subsection{Weight Initialization of Neural Networks}

Many prior works have studied the weight initialization for standard DNN training.
Xavier or Glorot initialization~\citep{Xavier}, adopted by popular deep learning libraries such as PyTorch~\citep{pytorch} and Tensorflow~\citep{abadi2016tensorflow} as the default initialization, aim to stabilize the magnitude of forward propagation and gradient backpropagation signals measured with variance.
It uses a uniform distribution or normal distribution to independently initialize each element in the weight matrix with a derived variance for the distribution.
\citet{He_2015_ICCV} derived an initialization that more accurately stabilizes the variance in ReLU networks.
\cite{saxe2013exact} proposed an orthogonal initialization which may lead to better learning dynamics.
Some other works also derived initializations for specific DNN structures~\citep{res_initialization,transformer_initialization}, and \citet{bhattacharya2020learnable,zhu2021gradinit} proposed to automatically learning initializations.
However, these initializations were designed for standard DNN training, while they can generally lead to exploded certified bounds for IBP training as we will show in this paper.
 
\subsection{Batch Normalization for DNN Training}
 
Batch normalization (BN)~\citep{ioffe2015batch} is originally proposed to improve DNN training by reducing interval covariate shift.
More recently, \citet{santurkar2018does} instead suggests that BN actually improves DNN training by smoothing the loss landscape without the necessity of reducing internal covariate shift, and BN can accelerate DNN training~\citep{van2017l2}. 
In this paper, we identify the extra benefit of using BN in IBP training.

\section{Methodology}

\subsection{Notations and Definitions}
\label{sec:definitions}

We focus on improving IBP training,
and we consider a commonly adopted $\ell_\infty$ perturbation setting in adversarial robustness on a $K$-way classification task.
For a DNN $f_\theta(\rvx)$ with clean input $\rvx$, there can be some perturbation $\delta$ satisfying $\|\delta\|_\infty\leq\eps$, and the actual perturbed input to the model is $\rvx+\delta$.
In robustness verification for achieving certified robustness, we verify whether 
\begin{equation}
[f_\theta(\rvx+\delta)]_y - [f_\theta(\rvx+\delta)]_i > 0, 
\quad 
\forall i,\delta 
~\st~ i \neq y,\, \|\delta\|_\infty\leq\eps,  
\label{eq:verify}
\end{equation}
holds true, where $[f_\theta(\rvx+\delta)]_i$ is the logits score for class $i$ and $y$ is the ground-truth. This is equivalent to verifying whether the DNN provably makes correct prediction for all input $\rvx+\delta~(\|\delta\|_\infty\leq\eps)$.
For network $f_\theta$, we assume that there are $m$ hidden affine layers (either convolutional or fully-connected layers) with ReLU activation.
We use $\rvh_i$ to denote the pre-activation output value of the $i$-th layer, and $\rvh_{i,j}$ denotes the $j$-th neuron in the $i$-th layer. 
We also use $\rvz_i=\text{ReLU}(\rvh_i)$ to denote the post-activation value. 
For a convolutional or fully-connected layer, we use $\rmW_i$ and $\rvb_i$ to denote its parameters, where $\rmW_i \in \mathbb{R}^{r_i \times n_i}, \rvb \in \mathbb{R}^{r_i}$, and $r_i$ and $n_i$ are called the ``fan-out'' and ``fan-in'' number of the layer respectively~\citep{he2015delving}.
This is straightforward for a fully-connected layer, and for a convolutional layer with kernel size $k$, $c_{\text{in}}$ input channels and $c_{\text{out}}$ output channels, we can still view the convolution as an affine transformation with  $n_i=k^2c_{\text{in}}$ and $r_i=c_{\text{out}}$.
In particular, we use $\rvh_0=\rvx+\delta$ to denote the input layer perturbed by $\delta$ ($\rvz_0$ is not applicable). 

For IBP~\citep{mirman2018differentiable,gowal2018effectiveness}, it computes and propagates lower and upper bound intervals layer by layer until the last layer or the verification objective.
For pre-activation $\rvh_i$, its interval bounds can be denoted as  $[\underline{\rvh}_i, \overline{\rvh}_i]$, where $ \underline{\rvh}_i \leq \rvh_i \leq \overline{\rvh}_i~(\forall \|\delta\|_\infty\leq\eps)$.
Similarly, there are also post-activation interval bounds $[\ul{\rvz}_i,\ol{\rvz}_i]$.
Finally \eqref{eq:verify} can be verified by checking the lower bound of $ [f_\theta(\rvx+\delta)]_y - [f_\theta(\rvx+\delta)]_i $. 
  
\subsection{Issues in Existing Certified Robust Training}
\label{sec:issues}

In this section, we analyze the issues in existing IBP training.
In particular, we identify two issues, including exploded bounds at initialization, and also the imbalance between ReLU activation states. 

\subsubsection{Exploded Bounds at Initialization}
\label{sec:exploded_bounds_at_init}

For simplicity, we assume the network has a feedforward architecture in this analysis, but the analysis can also be easily extended to other architectures.
For affine layer $\rvh_i=\rmW_i\rvz_{i-1}+\rvb_i$, the IBP bound computation is as follows:
\begin{equation}
\begin{aligned}
\underline{\rvh}_i = \rmW_{i,+}\underline{\rvz}_{i-1} + \rmW_{i,-}\overline{\rvz}_{i-1}+\rvb_i,\enskip
  \overline{\rvh}_i = \rmW_{i,+}\overline{\rvz}_{i-1} + \rmW_{i,-}\underline{\rvz}_{i-1}+\rvb_i,
\end{aligned}
\label{eq:ibp_affine}
\end{equation}
where $\rmW_{i,+}$ stands for retaining positive elements in $\rmW_i$ only while setting other elements to zero, and vice versa for $\rmW_{i,-}$.
$\rvh_i$ can be viewed as a function with the post-activation value of the previous layer $\rvz_i$ as input, denoted as $\rvh_i(\rvz_i) $.
In \eqref{eq:ibp_affine}, the IBP bounds guarantee that $\underline{\rvh}_i \leq \rvh_i(\rvz_i) \leq \overline{\rvh}_i~(\forall \underline{\rvz}_i\leq \rvz_i\leq \overline{\rvz}_i)$ for element-wise ``$\leq$''.

We then check the tightness of the interval bounds:
\begin{equation}
\Delta_i=\overline{\rvh}_i - \underline{\rvh}_i
=|\rmW_i|	(\overline{\rvz}_{i-1}-\underline{\rvz}_{i-1})=|\rmW_i|\delta_{i-1},
\end{equation}
where $\Delta_i$ denotes the gap between the upper and lower bounds, which can reflect the tightness of the bounds, and
$|\rmW_i|$ stands for taking the absolute value element-wise.
At initialization, we assume that each $\rmW_i$ independently follows a  distribution with zero mean and variance $\sigma_i^2$, and the distribution is symmetric about 0.
For a vector or matrix with independent elements following the same distribution, we use $\E(\cdot)$ to denote the expectation of this distribution.
We can view each element in vector $\Delta_i$ as a random variable that follows the same distribution, and we denote its expectation as $\E(\Delta_i)$, to measure the expected tightness at layer $i$.
As $\rmW_i$ and $\delta_{i-1}$ are independent, we have $\E(\Delta_i)=n_i\E(|\rmW_i|)\E(\delta_{i-1})$. 
Detailed in Appendix~\ref{apd:proof_delta},
we further have $ \E(\delta_i)=\E(\text{ReLU}(\overline{\rvh}_{i}) - \text{ReLU}(\underline{\rvh}_{i}))
     =\frac{1}{2} \E(\Delta_{i})$, and 
\begin{equation}
\E(\Delta_{i}) = \frac{n_i}{2} \E(|\rmW_i|)\E(\Delta_{i-1}).
\label{eq:e_grow}
\end{equation}
Empirically, we can estimate $\E(\Delta_i)$ given a batch of concrete data, by taking the mean, and we use $\hat{\E}(\Delta_{i})$ to denote the result of the empirical estimation.

We define a metric to characterize to what extent the certified bounds become looser, after propagating bounds from layer $i-1$ to layer $i$:
\begin{definition}
\label{def:diff_gain}
We define the difference gain when bounds are propagated from layer $i-1$ to layer $i$: 
\begin{equation}
\E(\Delta_i)/\E(\Delta_{i-1}) = \frac{n_i}{2} \E(|\rmW_i|).
\label{eq:diff_gain}	
\end{equation}
Bounds are considered to be stable if the difference gain $\E(\Delta_i)/\E(\Delta_{i-1})$ is close to 1.
\end{definition}
A large difference gain indicates exploded bounds, but it cannot be much smaller than 1 either to avoid signal vanishing in the model.
We find that weight initialization in prior works have large difference gain values especially for layers with larger $n_i$.
For example, for the widely used Xavier initialization~\citep{Xavier}, the difference gain is $\frac{1}{4}\sqrt{n_i}$, and it can be as large as 45.25 when $n_i=32768$ for a fully-connected layer in experiments. 
This indicates that certified bounds explode at initialization.
We illustrate the bound explosion in Figure~\ref{fig:E_delta}, and in Appendix~\ref{ap:illustration}, we list the difference gain of each existing initialization method in Table~\ref{tab:init}.
As a result, long warmup schedules are important in previous works, to gradually tighten certified bounds and ease training, but this is inefficient.

\subsubsection{Imbalanced ReLU Activation States}
\label{sec:imbalance_relu}
We show another issue in existing certified training, where the models have a bias towards \emph{inactive ReLU neurons}. 
Here ``inactive ReLU neurons'' are defined as neurons with non-positive pre-activation upper bounds ($\overline{\rvh}_{i,j} \!\leq\! 0$), i.e., they are always inactive regardless of input perturbations.
Similarly, \emph{active ReLU neurons} have non-negative pre-activation lower bounds ($\underline{\rvh}_{i,j} \!\geq\! 0$).
There are also \emph{unstable ReLU neurons} with uncertain activation states given different input perturbations ($\underline{\rvh}_{i,j} \!\leq\! 0 \!\leq\! \overline{\rvh}_{i,j}$).
In IBP training, inactive neurons have tighter bounds than active and unstable ones as shown in Figure~\ref{fig:relu_neuron_states} in Appendix~\ref{ap:exp}, and thus the optimization tends to push the neurons to be inactive. We show this imbalance ReLU status in Figure~\ref{fig:unbalanced_relu} (vanilla w/o BN), and it is more severe when the warmup is shorter as shown in Appendix~\ref{ap:relu_balance_short}.
Too many inactive neurons indicates that many neurons are essentially unused or dead, which will harm the model's capacity and block gradients as discussed by \cite{lu2019dying} on standard training.

\begin{figure}[tb]
\begin{minipage}{0.48\textwidth}
    \centering
    \includegraphics[width=\textwidth]{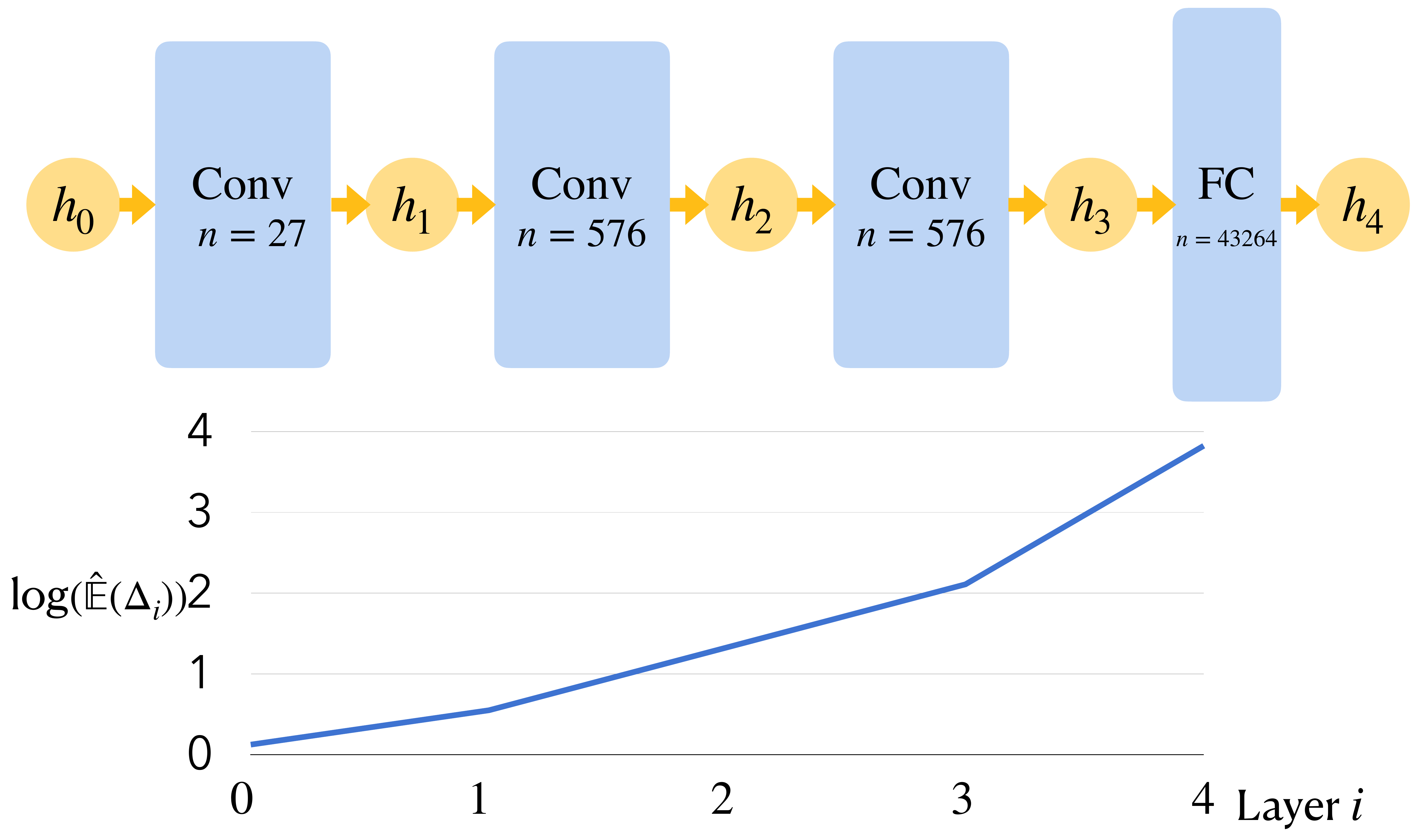}
    \caption{
    We show that certified bounds explode at initialization, in a simple untrained CNN (the classification layer is omitted) using Xavier initialization. We plot $\log \hat{\E}(\Delta_i)$ for each layer $i$.
    }
    \label{fig:E_delta}
    \end{minipage}
    \hfill
\begin{minipage}{0.48\textwidth}
\centering
\includegraphics[width=\textwidth]{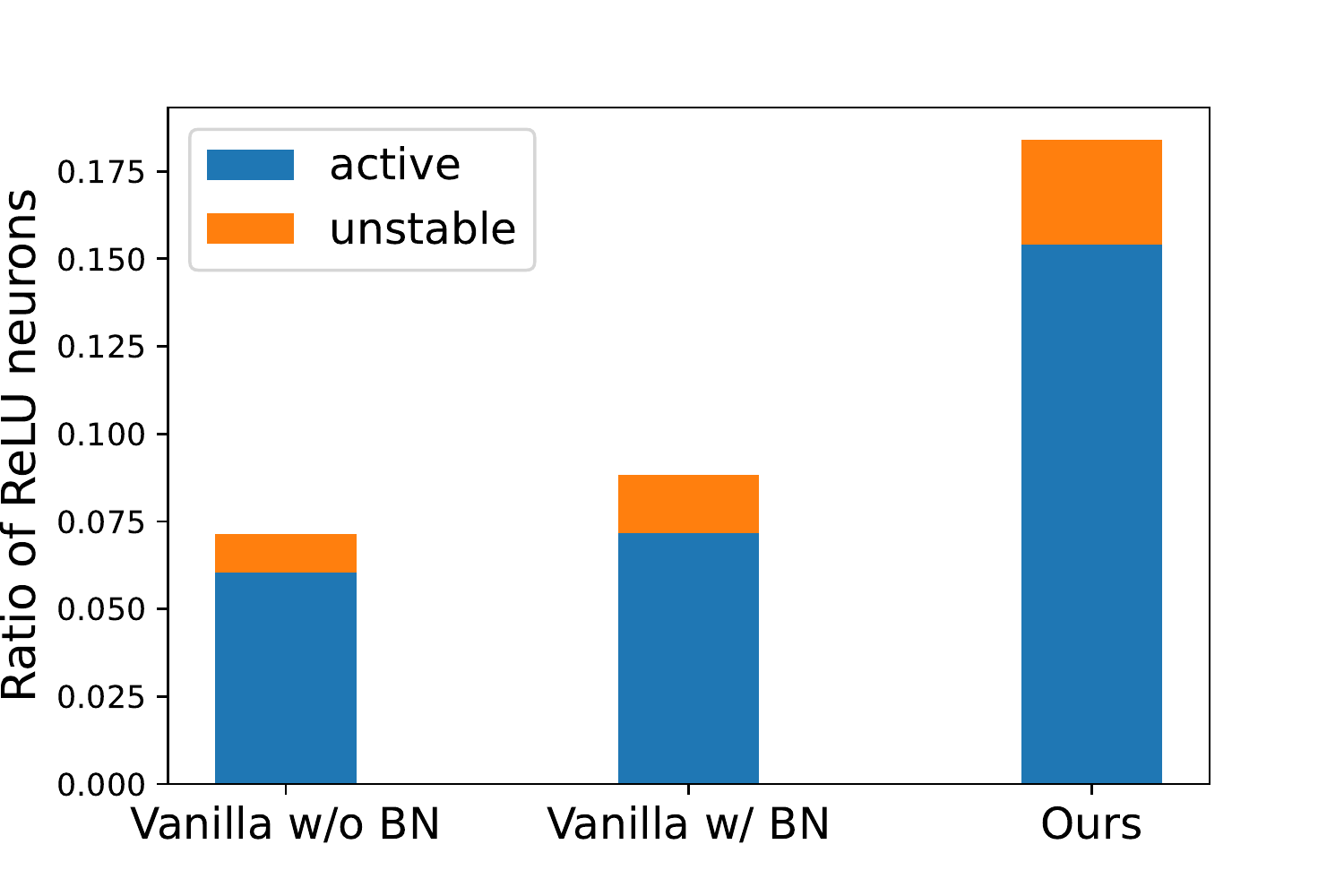}
\caption{Ratios of active and unstable ReLU neurons for CNN-7 on CIFAR-10 with different settings. The vanilla ones are not regularized, and ``vanilla (w/o BN)'' does not use BN either. 
}
\label{fig:unbalanced_relu}
\end{minipage}
\end{figure}

\subsection{The Proposed Method}
To address the aforementioned  issues, we propose our method in three parts:
1) We derive a new weight initialization for IBP training to stabilize the tightness of bounds at initialization;
2) We propose to fully add BN to  mitigate the ReLU imbalance and stabilize the variance of bounds, while models in prior works did not have BN for some or all the layers.
3) We further propose regularizations to explicitly stabilize the tightness and the balance of ReLU states during  warmup.

\subsubsection{IBP initialization}
\label{sec:init}
We propose a new \emph{IBP initialization} for IBP training.
Specifically, we independently initialize each element in $\rmW_i$ following a normal distribution $\gN(0,\sigma_i^2)$, and we aim to choose a value for $\sigma_i$ such that the \emph{difference gain} defined in \eqref{eq:diff_gain} is exactly 1.
When elements in $\rmW_i$ follow the normal distribution, we have $ \E(|\rmW_i|)=\sqrt{2/\pi} \sigma_i$, and thereby we take $\sigma_i=\frac{\sqrt{2\pi}}{n_i}$, which makes the difference gain $ \frac{n_i}{2}\E(|\rmW_i|)$ exactly 1. 
This initialization can further be calibrated for non-feedforward networks such as ResNet as we discuss in Appendix~\ref{apd:init_resnet}.

\subsubsection{Batch Normalization}
\label{sec:bn}
Batch normalization (BN)~\citep{ioffe2015batch} normalizes the input of each layer to a distribution with stable mean and variance. It can improve the optimization for DNN as shown in prior works for standard DNN training~\citep{ioffe2015batch,van2017l2,santurkar2018does}. 
In addition, for IBP training, BN can normalize the variance of bounds, and it can also improve the balance of ReLU activation states by shifting the center of upper and lower bounds to zero (before the additional linear transformation which comes after the normalization).
In prior certified training works~\citep{gowal2018effectiveness,zhang2019towards,xu2020automatic}, they only used BN for some layers in some models but not all layers, and they did not identify the benefit of BN in certified training. We empirically demonstrate that fully adding BN to each affine layer can significantly mitigate the imbalance ReLU issue and improve IBP training.
We follow the BN implementation by \citet{wong2018scaling,xu2020automatic} for certified training, where the shifting and scaling parameters are computed from unperturbed data. 

Note that our previous analysis on IBP initialization considers a network without BN. 
BN which rescales the output of each layer can still affect the tightness of IBP bounds, and the effect of IBP initialization may be weakened.
This is a limitation of the proposed initialization which could possibly be improved by considering the effect of BN in future work.
Nevertheless, in Appendix~\ref{ap:BN_IBP}, we empirically show that BN still does not cancel out the effect of IBP initialization.

\subsubsection{Warmup Regularization}
\label{sec:reg}

To further address the aforementioned two issues in Sec.~\ref{sec:issues},  and to explicitly stabilize the tightness of certified bounds and balance ReLU neuron states, 
we add two regularizers in the warmup stage of IBP training, 
The regularizers are principled and motivated by the two issues we discover.

\paragraph{Bound tightness regularizer}

Similar to the goal of stabilizing certified bounds at initialization, we also expect to keep the mean value of $\Delta_i$ in the current batch, $\hat{\E}(\Delta_{i})$, stable along the warmup.
Note that $\hat{\E}(\Delta_{i})$ is empirically computed from a concrete batch and different from the expectation $ \E(\Delta_i) $ at initialization 
In the initialization, we aim to make $\E(\Delta_i)\approx\E(\Delta_{i-1})$. 
Here, we relax the goal to making $\tau\hat{\E}(\Delta_i)\leq\hat{\E}(\Delta_0)$ with a configurable tolerance value $\tau\ (0\!<\!\tau\!\leq\!1)$, to balance the regularization power and the model capacity.
We add the following regularization term:
\begin{equation}
	\gL_{\text{tightness}}=\frac{1}{\tau m} \sum_{i=1}^m\text{ReLU}(\tau-\frac{\hat{\E}(\Delta_0)}{\hat{\E}(\Delta_i)}),
	\label{eq:l_tightness}
\end{equation}
where the training is penalized only when $\tau\hat{\E}(\Delta_i)>\hat{\E}(\Delta_0)$ due to the clipping effect by $\text{ReLU}(\cdot)$.

\paragraph{ReLU activation states balancing regularizer}

To balance ReLU activation states, we expect to balance the impact of active ReLU neurons and inactive neurons respectively.
Here, we consider the center of the interval bound, $\rvc_i=(\underline{\rvh}_i+\overline{\rvh}_i)/2$, and we model the impact as the contribution of each type of neurons to the mean and variance of the whole layer, i.e., $\hat{\E}(\rvc_i)$ and $\Var(\rvc_i)$ respectively.
Note that in the beginning almost all neurons are unstable, and gradually most neurons become either active or inactive.
Therefore, we add this regularizer only when there is at least one active neuron and one inactive neuron, which generally holds true unless at the training start.
We use $\alpha_i$ to denote the ratio between the contribution of the active neurons and inactive neurons respectively to $\hat{\E}(\rvc_i)$, and similarly we use $\beta_i$ to denote the ratio of contribution to $\Var(\rvc_i)$. They are computed as:
\begin{equation*}
\alpha_i=\frac{ \quad \sum_{j} \sI(\underline{\rvh}_{i,j}>0)\rvc_{i,j} }{ -\sum_{j} \sI(\overline{\rvh}_{i,j}<0)\rvc_{i,j} },\qquad
\beta_i=\frac{ \sum_{j} \sI(\underline{\rvh}_{i,j}>0) (\rvc_{i,j}-\hat{\E}(\rvc_i))^2 }{ \sum_{j} \sI(\overline{\rvh}_{i,j}<0)(\rvc_{i,j}-\hat{\E}(\rvc_i))^2 },
\end{equation*}
and in general $\alpha_i,\beta_i>0$.
We regard that the activation states are roughly balanced if $\alpha_i$ and $\beta_i$ are close to 1.
With the same aforementioned tolerance $\tau$, we expect to make $\tau\!\leq\!\alpha_i,\beta_i\!\leq \!1/\tau$, which is equivalent to making $\min(\alpha_i,1/\alpha_i)\geq \tau,\,\min(\beta_i,1/\beta_i) \geq\tau$.
Thereby we design the following regularization term:
\begin{equation}
\gL_{\text{relu}} = \frac{1}{\tau m}\sum_{i=1}^m\bigg(
\text{ReLU}(\tau-\min(\alpha_i,\frac{1}{\alpha_i}))+
\text{ReLU}(\tau-\min(\beta_i,\frac{1}{\beta_i}))
\bigg).
\label{eq:l_relu}
\end{equation}

\subsection{Training Objectives}

Certified robust training solves the robust optimization problem as \eqref{eq:minmax_opt}, and when the inner maximization is verifiably solved, the base training objective without regularization is:
\begin{equation}
\gL_{\text{rob}} = \overline{L}(f_\theta,\rvx, y, \eps),\quad
\text{where} \enskip\overline{L}(f_\theta,\rvx, y, \eps)\geq\max_{\|\delta\|_\infty\leq\eps} L(f_\theta(\rvx+\delta),y),
	\label{eq:ord_obj}
\end{equation}
such that $\overline{L}(f_\theta,\rvx,y,\eps)$ is an upper bound of $L(f_\theta(\rvx+\delta),y)$ given by a robustness verifier, e.g., IBP.
In our proposed method, we first initialize the parameters with our IBP initialization,
and then we perform a \emph{short} warmup with gradually increasing $\eps~(0\leq\eps\leq\epst)$, where $\epst$ stands for the target perturbation radius that is usually equal to or slightly larger than the maximum perturbation radius used for test.
Our training objective $\gL$ combines the ordinary objective \eqref{eq:ord_obj} and the proposed regularizers:
\begin{equation}
	\mathcal{L}=\gL_{\text{rob}} + \lambda ( \gL_{\text{tightness}} + \gL_{\text{relu}}  ),
	\label{eq:l_full}
\end{equation}
where $\lambda$ is for balancing the regularizers and the original $\gL_{\text{rob}}$ loss.
For simplicity and efficiency, we use IBP to compute the bounds in $\gL_{\text{rob}}$ and the regularizers.
During warmup, we also gradually decrease $\lambda$ from $\lambda_0$ to 0 as $\eps$ grows, where 
$ \lambda=\lambda_0(1-\eps/\epst)$.
After warmup, we only use $\gL=\gL_{\text{rob}}$ for final training with $\epst$.
Note that in the regularizers, the value of each $\text{ReLU}(\cdot)$ term has the same range $[0,\tau]$, and thus in \eqref{eq:l_full} we directly sum up them without weighing them for simplicity. In test, we still only use pure IBP bounds  without any other tighter method.

\section{Experiments}

\label{sec:exp}

In the experiments, we demonstrate the effectiveness of our proposed method for training certifiably robust neural networks more efficiently while achieving better or comparable verified errors.

\subsection{Settings}

\label{sec:exp_settings}

We adopt three datasets, MNIST~\citep{mnist}, CIFAR-10~\citep{cifar} and TinyImageNet~\citep{le2015tiny}.
Following \citet{xu2020automatic}, we consider three model architectures: a 7-layer feedforward convolutional network (CNN-7), Wide-ResNet~\citep{zagoruyko2016wide} and ResNeXt~\citep{xie2017aggregated}.
According our discussion in Sec.~\ref{sec:bn}, we also modify the models to fully add a BN after every convolutional or fully-connected layer.
For target perturbation radii, we mainly use $\epst=0.4$ for MNIST, $\epst=8/255$ for CIFAR-10, and $\epst=1/255$ for TinyImageNet, following prior works, and we provide results on other perturbation radii in Appendix~\ref{ap:multi_eps}.
We provide more implementation details in Appendix~\ref{ap:imp}.
We mainly compare with the following SOTA baselines on all the settings (note that in our main results, we also make these baselines use models with full BNs unless otherwise indicated):

\begin{itemize}[leftmargin=*]
\item Vanilla IBP~\citep{gowal2018effectiveness} with existing initialization and no warmup regularizer. We use the default Xavier initialization in PyTorch, and we find that  orthogonal initialization originally used by \cite{gowal2018effectiveness} does not improve the performance here.

\item CROWN-IBP~\citep{zhang2019towards} with linear relaxation bounds by CROWN~\citep{zhang2018efficient} during warmup. We use the generalized and accelerated version with loss fusion by \citet{xu2020automatic}, while the original version is $O(K)$ (the number of classes) more costly.
During the warmup, it combines bounds by IBP and linear relaxation with weight $\eps/\epst$ and $(1-\eps/\epst)$ respectively. 
\end{itemize}

\subsection{Certified Robust Training with Short Warmup}

\begin{table*}[!ht]
  \centering
  \caption{Standard and verified error rates (\%) of models trained with different methods respectively on MNIST ($\epst\!=\!0.4$) and CIFAR-10 ($\epst\!=\!8/255$). 
  Schedule is represented as the total number of epochs and the number of epochs in each of the three phases with $\eps=0$, increasing $\eps\in(0,\epst)$ and final $\eps=\epst$ respectively.
  We report the mean and standard deviation of the results on 5 repeats for CNN-7 and 3 repeats for Wide-ResNet and ResNeXt respectively. All models include BN after every layer (see Sec.~\ref{sec:bn}).
  We also report the best run in ``Ours (best)'' since main results in prior works did not have repeats.
  Literature results with the ``$\dagger$'' mark are concurrent works.
}
  \label{tab:main}
  \footnotesize
  \adjustbox{max width=\textwidth}{
  \begin{threeparttable}
  	\begin{tabular}{c|c|c|cc|cc|cc}
    \toprule[1pt]
    
    \multirow{2}{*}{Dataset} & Schedule & \multirow{2}{*}{Method}  & \multicolumn{2}{c|}{CNN-7 (with full BN)} & \multicolumn{2}{c}{Wide-ResNet (with full BN)} & \multicolumn{2}{c}{ResNeXt (with full BN)}\\
    & (epochs)  & & Standard & Verified & Standard & Verified & Standard & Verified\\
	\midrule[1pt]

\multirow{9}{*}{MNIST} &
\multirow{4}{*}{70 (0+20+50)} &
Vanilla IBP & 2.59 $\pm$ 0.06 &   12.03 $\pm$ 0.09 & 3.18 $\pm$ 0.05 &   12.93 $\pm$ 0.17 & 4.09 $\pm$ 0.46 &   15.36 $\pm$ 0.94  \\
& & CROWN-IBP~\tnote{a} & 2.75 $\pm$ 0.12 &   12.04 $\pm$ 0.22 & 3.39 $\pm$ 0.05 &   13.10 $\pm$ 0.15 & 4.22 $\pm$ 0.53 &   15.24 $\pm$ 0.78  \\
& & Ours & \textbf{2.33 $\pm$ 0.08} & \textbf{11.03 $\pm$ 0.13} & \textbf{2.77 $\pm$ 0.02} & \textbf{11.76 $\pm$ 0.07} & \textbf{3.22 $\pm$ 0.08} & \textbf{13.43 $\pm$ 0.17}  \\
& & Ours (best) & \textbf{2.20} & \textbf{10.82} & 2.75 & 11.69 & 3.17 & 13.20  \\

\cline{2-9}
& \multicolumn{3}{c}{Literature results} & \multicolumn{2}{c}{Warmup} & \multicolumn{1}{c}{Total (epochs) } & Standard & Verified\\
\cline{2-9}
& \multicolumn{3}{c}{\citet{gowal2018effectiveness}} & \multicolumn{2}{c}{(2K+10K) steps} & \multicolumn{1}{c}{100} & 1.66 & 15.01~\tnote{b} \\
& \multicolumn{3}{c}{\citet{zhang2019towards}} & \multicolumn{2}{c}{$(9+51)$ epochs} & \multicolumn{1}{c}{200} & 2.17 & 12.06 \\
\cline{2-9}
& \multicolumn{3}{c}{$^\dagger$IBP+ParamRamp~\citep{lyu2021evaluating}~\tnote{e}} & \multicolumn{2}{c}{$(9+51)$ epochs} & \multicolumn{1}{c}{200} & 2.16 & 10.88 \\
& \multicolumn{3}{c}{$^\dagger$CROWN-IBP+ParamRamp~\citep{lyu2021evaluating}~\tnote{e}} & \multicolumn{2}{c}{$(9+51)$ epochs} & \multicolumn{1}{c}{200} & 2.36 & 10.61 \\

\midrule[1pt]

\multirow{15}{*}{CIFAR-10} &
\multirow{3}{*}{70 (1+20+49)} &
Vanilla IBP & 58.72 $\pm$ 0.27 &   69.88 $\pm$ 0.10 & 58.85 $\pm$ 0.22 &   69.77 $\pm$ 0.32 & 60.10 $\pm$ 0.27 &   71.19 $\pm$ 0.21  \\
& & CROWN-IBP~\tnote{a} & 63.19 $\pm$ 0.36 &   71.29 $\pm$ 0.19 & 62.76 $\pm$ 0.23 &   71.82 $\pm$ 0.30 & 64.75 $\pm$ 0.50 &   72.50 $\pm$ 0.20  \\
& & Ours & \textbf{56.64 $\pm$ 0.48} & \textbf{68.81 $\pm$ 0.24} & \textbf{56.74 $\pm$ 0.40} & \textbf{68.71 $\pm$ 0.29} & \textbf{59.33 $\pm$ 0.86} & \textbf{70.62 $\pm$ 0.59}  \\
\cline{2-9} &
\multirow{4}{*}{160 (1+80+79)} &
Vanilla IBP & 53.80 $\pm$ 0.71 &   67.01 $\pm$ 0.29 & 54.31 $\pm$ 0.46 &   67.45 $\pm$ 0.21 & 55.23 $\pm$ 0.12 &   68.28 $\pm$ 0.15  \\
& & CROWN-IBP~\tnote{a} & 58.76 $\pm$ 0.76 &   69.67 $\pm$ 0.38 & 60.39 $\pm$ 0.33 &   70.07 $\pm$ 0.42 & 61.08 $\pm$ 0.35 &   71.26 $\pm$ 0.11  \\
& & Ours & \textbf{51.72 $\pm$ 0.40} & \textbf{65.58 $\pm$ 0.32} & \textbf{51.95 $\pm$ 0.27} & \textbf{65.91 $\pm$ 0.14} & \textbf{53.68 $\pm$ 0.33} & \textbf{66.91 $\pm$ 0.40}  \\
& & Ours (best) & \textbf{51.06} & \textbf{65.03} & 
51.63 & 65.72 & 
53.38 & 66.41\\

\cline{2-9}
& \multicolumn{3}{c}{Literature results} & \multicolumn{2}{c}{Warmup} & \multicolumn{1}{c}{Total (epochs) } & Standard & Verified\\
\cline{2-9}
& \multicolumn{3}{c}{\citet{gowal2018effectiveness}} & \multicolumn{2}{c}{(5K+50K) steps} & \multicolumn{1}{c}{3,200} & 50.51 & 68.44~\tnote{c} \\
& \multicolumn{3}{c}{\citet{zhang2019towards}} & \multicolumn{2}{c}{$(320+1600)$ epochs} & \multicolumn{1}{c}{3,200} & 54.02 &66.94 \\
& \multicolumn{3}{c}{\citet{balunovic2020adversarial}} & \multicolumn{2}{c}{N/A \tnote{d}} & \multicolumn{1}{c}{800} & 48.3 & 72.5 \\
& \multicolumn{3}{c}{\citet{xu2020automatic}} & \multicolumn{2}{c}{$(100+800)$ epochs} & \multicolumn{1}{c}{2,000} & 53.71 & 66.62 \\
\cline{2-9}
& \multicolumn{3}{c}{$^\dagger$IBP+ParamRamp~\citep{lyu2021evaluating}~\tnote{e}} & \multicolumn{2}{c}{$(320+1600)$ epochs} & \multicolumn{1}{c}{3,200} &55.28& 67.09 \\
& \multicolumn{3}{c}{$^\dagger$CROWN-IBP+ParamRamp~\citep{lyu2021evaluating}~\tnote{e}} & \multicolumn{2}{c}{$(320+1600)$ epochs} & \multicolumn{1}{c}{3,200} &51.94&65.08\\
& \multicolumn{3}{c}{$^\dagger$$\ell_\infty$-dist net (other architecture)~\citep{zhang2021towards}~\tnote{f} } & \multicolumn{2}{c}{N/A~\tnote{f}} & \multicolumn{1}{c}{800} & 48.32 & 64.90\\%

	\bottomrule[1pt]
  \end{tabular}

  \begin{tablenotes}\large
  \item[a] CROWN-IBP here follows  \citet{xu2020automatic} with loss fusion for efficiency, but we found it does not perform well with a short training schedule under our settings and usually requires a longer schedule to achieve good results.
  \item[b] Some test results in \citet{gowal2018effectiveness} are obtained with costly mixed integer programming (MIP) and linear programming (LP); we take IBP verified errors for fair comparison following \citet{zhang2019towards}.
  \item[c] Additional PGD adversarial training was involved for this result, according to \citet{zhang2019towards}. 
  \item[d] \citet{balunovic2020adversarial} used a different training scheme and train the network layer by layer.
  \item[e] \citet{lyu2021evaluating} use IBP-based and CROWN-IBP-based training respectively with their parameterized activation, and they use a tighter linear bound propagation method for testing instead of IBP. 
  \item[f]
  \citet{zhang2021towards} use a very different model architecture with $\ell_\infty$ distance neurons rather than traditional DNNs, but still need a long schedule on both $\eps$ and $\ell_p$ norm where $p$ is gradually increased until $\infty$.
  \end{tablenotes}
  \end{threeparttable}
  }
\end{table*}

\begin{table*}[ht]
    \centering
    \caption{
    Standard and verified error rates (\%) on TinyImageNet ($\eps_t=1/255$). The best result in literature~\citep{xu2020automatic} has a standard error of 72.18\% and verified error of 84.14\% using 800 epochs. We achieve 82.36\% verified error using only 80 epochs.
  }
  
  \begin{threeparttable}
  \adjustbox{max width=.85\textwidth}{
    \begin{tabular}{c|c|cc|cc|cc}
    \toprule
         \multirow{2}{*}{Model (with full BN)} &{Schedule} &  \multicolumn{2}{c|}{Vanilla IBP} & \multicolumn{2}{c|}{CROWN-IBP}& \multicolumn{2}{c}{Ours} \\
         & (epochs) & Standard & Verified & Standard & Verified & Standard & Verified  \\
         \midrule
         \multirow{2}{*}{CNN-7} & 80 (1+10+69) &   75.50 & 82.92 & 76.00 & 82.81 & 75.20 & \textbf{82.45}  \\
         & 80 (1+20+59) & 74.68 & 82.84 & 76.27 & 83.35 & 74.29 & \textbf{82.36}\\
         \cline{1-8}
         \multirow{2}{*}{Wide-ResNet~\tnote{a}} & 80 (1+10+69) & 75.89 & 83.00 & 75.85 & 83.65& 74.90 & \textbf{82.49} \\
         & 80 (1+20+59) & 75.65 & 83.17 & 75.95 & 83.08 & 74.59 & \textbf{82.75} \\
         \cline{1-8}
         \multirow{2}{*}{ResNeXt} &  80(1+10+69) & 82.39 & 87.15 &  85.47& 89.11 & 80.20 & \textbf{85.77} \\
         & 80 (1+20+59) & 81.72 & 87.10 & 80.81 & 86.43& 78.91 & \textbf{85.78} \\
    \bottomrule
    \end{tabular}}
    
\begin{tablenotes}\footnotesize
\item[a] The Wide-ResNet model used here is 5 times smaller than the one used in \cite{xu2020automatic} to save training \\ time. Additionally, we include BN after every layer in all models (see Section~\ref{sec:bn}).
\end{tablenotes}
 
  \end{threeparttable}
\label{tab:tiny_imagenet}
\end{table*}

We conduct certified robust training using relatively short warmup schedules to demonstrate the effectiveness of our proposed techniques for fast training.
We show the results in Table~\ref{tab:main} for MNIST, CIFAR-10 and Table~\ref{tab:tiny_imagenet} for TinyImageNet.
Compared to Vanilla IBP and CROWN-IBP, our improved IBP training consistently achieves lower standard errors and verified errors under same schedules respectively, where BN is added to the models for all these three training methods.
We find that CROWN-IBP with loss fusion~\citep{xu2020automatic} tends to require a larger number of epochs to obtain good results and it sometimes underperform Vanilla IBP under short schedules, but disabling loss fusion can make it much more costly and unscalable.
In terms of the best results, 
we achieve verified error $10.82\%$ on MNIST $\epst=0.4$, $65.03\%$ on CIFAR-10 $\epst=8/255$, and $82.36\%$ on TinyImageNet $\epst=1/255$, 
which makes a notable improvement over literature SOTA~\citep{gowal2018effectiveness,xu2020automatic} that used long training schedules. 
Compared to concurrent works~\citep{lyu2021evaluating,zhang2021towards} which use different improvement techniques, we have comparable verified errors, but they still need long training schedules.
For reference, we tried  \citet{zhang2021towards} which used a different architecture with ``$\ell_\infty$ distance neurons'' rather than convolution-based DNNs. On CIFAR-10 using 160 total epochs by reducing their training schedule proportionally, their verified error is 68.44\% which is much higher than ours.
Overall, the results demonstrate that our improved IBP training is effective for more efficient certified robust training with a shorter warmup. 

\subsection{Comparison on Training Cost}
\label{sec:time}

\begin{table*}[ht]
    \centering
    \caption{
    Comparison of estimated time cost (seconds), for CNN-7 on CIFAR-10.
    We report the total time, and also the per-epoch time during three training phases of $\epsilon$ schedule for methods with a short warmup.
    Literature results with the ``$\dagger$'' mark are considered as concurrent.
   }
   \label{tab:time}
   \footnotesize
   \begin{threeparttable}
   
  \adjustbox{max width=\textwidth}{

    \begin{tabular}{c|cc|cccc}
    \toprule

   \multicolumn{2}{c}{\multirow{2}{*}{Method}} & \multirow{2}{*}{Epochs} & \multicolumn{3}{c}{Epoch time in each phase (s)} & \multirow{2}{*}{Total time (s)}  \\
    
    \multicolumn{2}{c}{} & & $0$ & $(0,\epst)$ & $\epst$ \\
    \midrule

    \multirow{5}{*}{Literature Results}& IBP~\citep{gowal2018effectiveness} & 3200 & \multicolumn{3}{c}{\multirow{3}{*}{-}} & $40496\times 4$~\tnote{a}\\
    & CROWN-IBP (w/o loss fusion)~\citep{zhang2019towards} & 3200 & & & &  $91288\times 4$~\tnote{a}\\
    & CROWN-IBP~\citep{xu2020automatic} & 2000 & & & &   $52362\times 4$~\tnote{a}\\
	\cline{2-7}
	& $^\dagger$IBP+ParamRamp~\citep{lyu2021evaluating} & 3200 & \multicolumn{3}{c}{\multirow{2}{*}{-}} & $40496\times 4\times 1.09$~\tnote{b} \\
	& $^\dagger$CROWN-IBP+ParamRamp~\citep{lyu2021evaluating} & 3200 & & &&  $91288\times 4\times 1.51$~\tnote{b}\\
    \cline{1-7}
    \multirow{3}{*}{Short Warmup} & Vanilla IBP & 160  
    & 30.0 & 54.8 & 54.8 & 8747.9\\ 
    &CROWN-IBP & 160 &
    30.0 & 78.5 & 54.8 & 10641.3\\
    &Ours & 160
    & 64.0 & 64.0 & 54.8 & 9512.3\\
	\hline
    \end{tabular}
    }
    \end{threeparttable}

    \begin{flushleft}
    $^\text{a}$ 4 GPUs were used and their models are slightly different (we add BN after every layer).\\
    $^\text{b}$ The factors $1.09$ and $1.51$ are the overhead of their method reported by~\citep{lyu2021evaluating} when combining with IBP or CROWN-IBP.
    \end{flushleft}

    \label{tab:cost}
    
    \end{table*}

We compare the training cost using a single NVIDIA RTX 2080 Ti GPU.
For methods using short warmup, we measure the per-epoch time cost during three different phases, namely $\eps\!=\!0$, $0\!<\!\eps\!<\!\epst$, and $\eps=\epst$, and we then estimate the total training time according to the schedule. We use gradient accumulation wherever needed to fit the training into the memory of a single GPU.
We also compare with total time cost with literature methods using long schedules.
We show the results of CNN-7 for CIFAR-10 in Table~\ref{tab:time}, and other settings in Appendix~\ref{ap:time}.
For $\eps=0$, Vanilla IBP and CROWN-IBP use regular training while we compute IBP bounds for regularization and have a small overhead, but this phase is extremely short (no more than 1 epoch here). For $0\!<\!\eps\!<\!\epst$, our method has a small overhead on regularizers compared to Vanilla IBP, while CROWN-IBP using linear relaxation can be more costly. For $\eps=\epst$, all the three methods use the same pure IBP. 

For total time on CIFAR-10 with the same 160-epoch schedule, we only have a small overhead of around $9\%\!\sim\!13\%$ compared to Vanilla IBP and the cost is still around $12\%\!\sim\!23\%$ lower than  CROWN-IBP, while we achieve lower verified errors than the baselines under such short warmup schedules (see Table~\ref{tab:main}).
And importantly, compared to literature using long training schedules, we significantly reduce the number of training epochs and the total training time (e.g., \citet{xu2020automatic} is around $20\times$ more costly than ours in total).

\subsection{Ablation Study and Discussions}
\label{sec:ablation}
\label{sec:discussion}

\begin{table}[t]
  \centering
    \caption{
    Standard error and verified error rates (\%) in the ablation study with CNN-7 on CIFAR-10. ``BN-Conv'' stands for BN after each convolutions, and ``BN-FC'' stands for BN after the hidden fully-connected layer.
    ``$\checkmark$'' means that the  component is enabled, and ``$\times$'' means that the component is disabled. We repeat each setting for 5 times and report the mean and standard deviation.
   }
  \footnotesize
  \adjustbox{max width=\textwidth}{
  \begin{tabular}{ccccc|cc|cc}
    \toprule[1pt]
    
   BN-Conv & BN-FC & IBP Initialization & $\gL_{\text{tightness}}$ & $\gL_{\text{relu}}$ & \multicolumn{2}{c|}{70 (1+20+49) epochs} & \multicolumn{2}{c}{160 (1+80+79) epochs}  \\
   & & & & & Standard & Verified & Standard & Verified\\
   \midrule

$\times$ & $\times$ & $\times$ & $\times$ & $\times$ & 59.33$\pm$0.70 & 70.18$\pm$0.18& 57.08$\pm$0.29 & 69.43$\pm$0.28\\ 

$\checkmark$ & $\times$ & $\times$ & $\times$ & $\times$ & 61.95$\pm$0.80 & 71.12$\pm$0.42& 57.21$\pm$0.65 & 69.21$\pm$0.30\\ 

$\checkmark$ & $\checkmark$ & $\times$ & $\times$ & $\times$ & 58.72$\pm$0.27 & 69.88$\pm$0.10& 53.80$\pm$0.71 & 67.01$\pm$0.29\\ 

$\checkmark$ & $\checkmark$ & $\checkmark$ & $\times$ & $\times$ & 58.93$\pm$0.29 & 69.60$\pm$0.35& 54.59$\pm$0.64 & 67.63$\pm$0.34\\ 

$\checkmark$ & $\checkmark$ & $\checkmark$ & $\checkmark$ & $\times$  & 56.76$\pm$0.38 & 68.96$\pm$0.49& 53.08$\pm$0.26 & 66.74$\pm$0.20\\ 

$\checkmark$ & $\checkmark$ & $\checkmark$ & $\times$ & $\checkmark$  & 58.49$\pm$0.42 & 69.38$\pm$0.23& 53.29$\pm$0.76 & 66.46$\pm$0.44\\ 

$\checkmark$ & $\checkmark$ & $\times$ & $\checkmark$ & $\checkmark$  & 58.79$\pm$0.40 & 69.29$\pm$0.28 & 52.45$\pm$0.34 & 66.34$\pm$0.38\\ 

$\checkmark$ & $\checkmark$ & $\checkmark$ & $\checkmark$ & $\checkmark$ & \textbf{56.64$\pm$0.48} & \textbf{68.81$\pm$0.24} & \textbf{51.72$\pm$0.40} & \textbf{65.58$\pm$0.32}\\ 

    \bottomrule[1pt]
  \end{tabular}
    }
  \label{table:ablation}
\end{table}

In this section, we empirically verify whether each part of our modification contributes to the improvement and whether they behave as we expect.
We conduct an ablation study and  also plot the curve of the regularization terms to reflect the bound tightness and ReLU balance during training.

We use CIFAR-10 with the currently best CNN-7 model under the ``$1+20$'' and ``$1+80$'' warmup schedules as used in Table~\ref{tab:main}.
We report the results in Table~\ref{table:ablation}. 
The first three rows show that fully adding BN improves the training when vanilla IBP is used, and it is important to add BN for the fully-connected layer, which was missed in prior works.
Based on the improved model structure, adding both IBP initialization and warmup regularization further improves the performance, and removing either of these parts leads to a degraded performance.

We notice that adding IBP initialization without warmup regularization may not improve the verified error. A factor is that IBP initialization can reduce the variance of the outputs (see Appendix~\ref{apd:bound_variance}), and it may harm the training during the early warmup, when $\epsilon$ is small and certified training is close to standard training. Also, the effect of initialization can be weakened when $\eps$ is much smaller than $\epst$. 
But the warmup regularization can continue to tighten the bounds, and the IBP initialization can benefit the optimization for the tightness regularizer. 
Nevertheless, IBP initialization is more beneficial for deep models where the exploded bound issue is more severe (see Appendix~\ref{ap:resnext_explosion}). 

It is also important to fully add BN to make the warmup regularization work well.
BN can normalize the variance of the layers, so the tightness regularizer can more effectively tighten certified bounds w.r.t. the stable variance; otherwise the the training may trivially optimize tightness regularizer by making the magnitude of the network output small.

\begin{figure}[t]
\centering
\begin{minipage}[t]{0.48\textwidth}
    \centering
    \includegraphics[width=.95\textwidth]{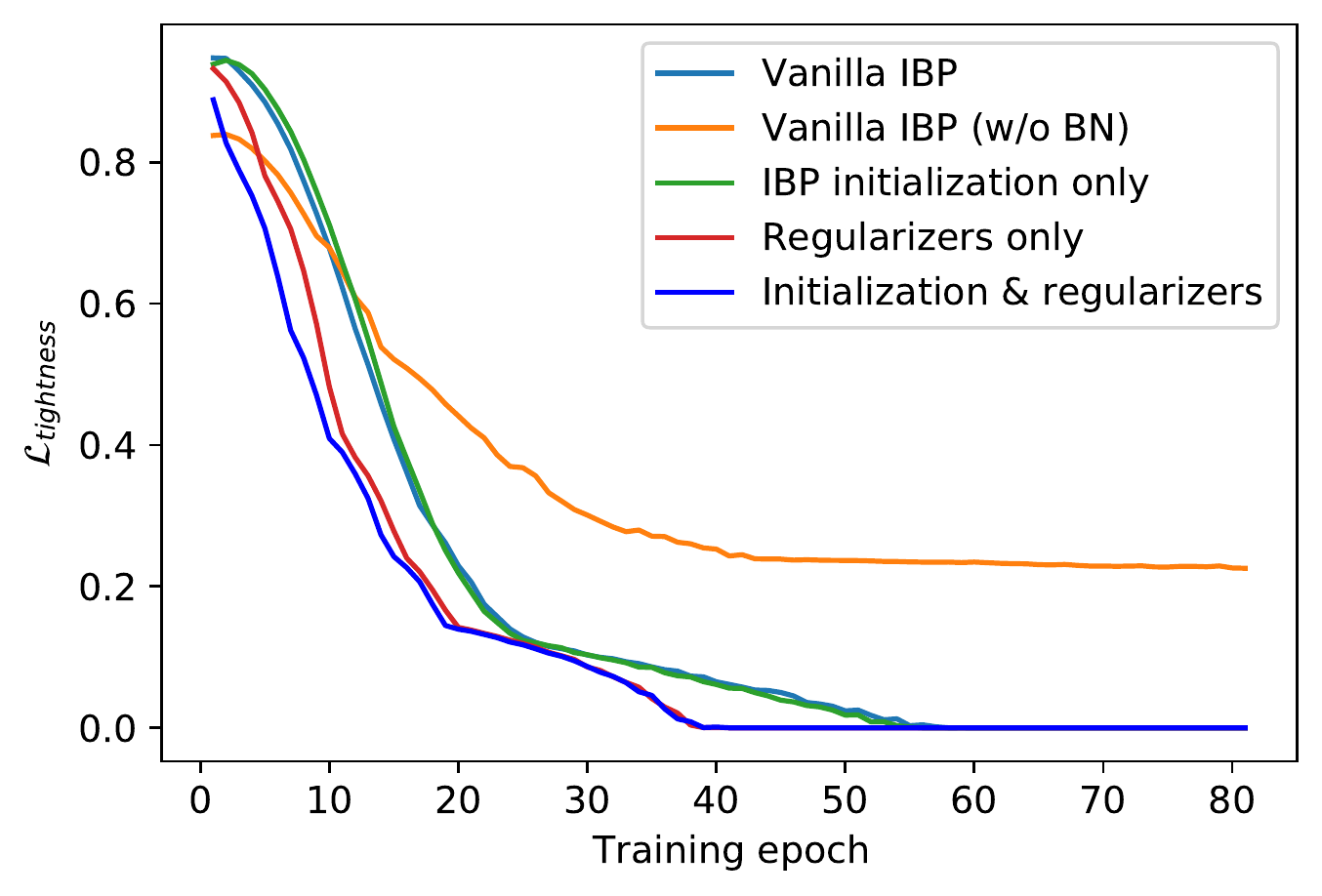}
	\caption{$\gL_{\text{tightness}}$ during warmup. $\gL_{\text{tightness}}$ is optimized only for ``regularizers only'' and ``initialization \& regularizers'' setting, and BN is fully added to every layer except for ``Vanilla IBP (w/o BN)''.}
	\label{fig:l_tightness}
    \end{minipage}\hfill
\begin{minipage}[t]{0.48\textwidth}
\centering
\includegraphics[width=.95\textwidth]{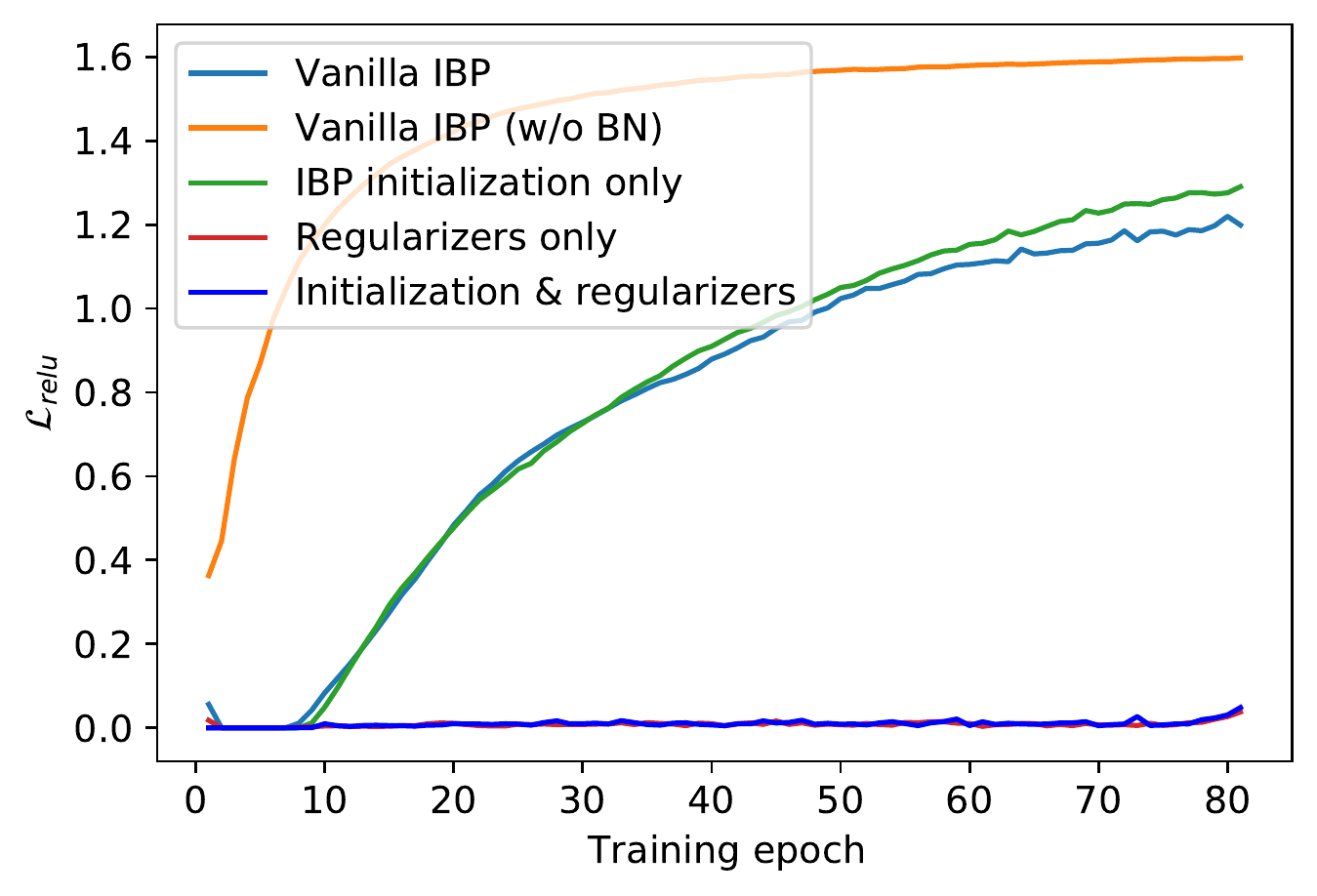}
	\caption{$\gL_{\text{relu}}$ during warmup, under same setting as in Figure~\ref{fig:l_tightness}.}
	\label{fig:l_act}
\end{minipage}
\end{figure}

Finally, we also plot the training curves of the regularizers to confirm if the regularizers are effectively optimized, so that the bound tightness and ReLU balance are indeed improved. Note that for the settings without regularizers, we only plot but not optimize the regularizers.
In Figure~\ref{fig:l_tightness}, we plot $\mathcal{L}_\text{tightness}$. By using the regularization in training, $\gL_{\text{tightness}}$ descends faster, and further adding the IBP initialization leads to even faster descent during the early epochs. 
In Figure~\ref{fig:l_act}, we show that the $\mathcal{L}_\text{relu}$ is indeed under control when we optimize it, while $\mathcal{L}_\text{relu}$ could gradually grow larger when the it is not added in training. Notably, when BN is removed and the regularization term is not optimized (Vanilla IBP (w/o BN)), $\gL_{\text{relu}}$ becomes extremely large in later epochs, and $\gL_{\text{tightness}}$ is also large in the end, which suggests that the training is hampered.
\section{Conclusion}

In this paper, we identify two issues in existing certified robust training methods regarding exploded bounds and imbalanced ReLU neuron states. To address these issues based on IBP training, we propose an IBP initialization and warmup regularization, and we also identify the benefit of fully adding BN. With our improvements, we demonstrate that we are able to achieve better verified errors using much shorter warmup and training schedules compared to literatures under the same convolution-based network architecture, for fast certified robust training.
\section*{Acknowledgement}
This work is supported in part by NSF under IIS-1901527, IIS-2008173, IIS-2048280 and by Army Research Laboratory under agreement number W911NF-20-2-0158.
\clearpage
\bibliography{main}
\bibliographystyle{arxiv}
\clearpage
\appendix
\section{Supplementary Illustrations for Motivation and Methodology}

\label{ap:illustration}

\subsection{List of Initialization Methods in Prior Works}

\begin{table}[ht]
	\caption{List of several weight initialization methods and their \emph{difference gain}.
	We show each difference gain in both closed form, and also empirical values when $n_i\in\{27,576,1152,32768\}$ for a 7-layer CNN model (without BN). 
	The concrete values are obtained by computing the mean of 100 random trials respectively.
	For orthogonal initialization, obtaining a closed form of difference gain is non-trivial so we omit its closed-form result, but it has large difference gains under empirical measurements.}
	\label{tab:init}
	\centering
	\adjustbox{max width=\textwidth}{
	\begin{tabular}{ccccccc}
	\toprule[1pt]
	\multirow{2}{*}{Method}  & \multirow{2}{*}{Adopted by} & \multicolumn{5}{c}{Difference Gain}\\
	& & Closed form & $n_i=27$ & $n_i=576$ & $n_i=1152$ & $ n_i=32768$ \\
	\midrule[1pt]
	Xavier (uniform)~\citep{Xavier} & \citet{zhang2019towards,xu2020automatic} & $ \frac{1}{4} \sqrt{n_i}$ &  1.30 & 6.00 & 8.48 & 45.25 \\
	Xavier (Gaussian)~\citep{Xavier} & -  & $ \sqrt{\frac{1}{2\pi}}\sqrt{n_i}$ &  2.07 & 9.57 & 13.54 & 72.2 \\
	Kaiming (uniform)~\citep{he2015delving}  & - & $\frac{\sqrt{3}}{4} \sqrt{n_i}$ & 3.20 &14.70& 20.77 &110.85\\
	Kaiming (Gaussian)~\citep{he2015delving} & - &   $ \sqrt{\frac{1}{\pi}}\sqrt{n_i} $ & 2.93 & 13.54 & 19.15 & 102.13 \\ 
	Orthogonal~\citep{saxe2013exact} & \citet{gowal2018effectiveness} & - & 2.09 &9.58 &13.54& 72.22 \\
	\hline
	IBP Initialization & This work & $ 1$ & 1.01 &1.00 &1.00& 1.00 \\
	\bottomrule[1pt]
	\end{tabular}}
\end{table}

In Table~\ref{tab:init}, we list several weight initialization methods and their corresponding difference gain  (see Def.~\ref{def:diff_gain}).
Prior weight initialization methods lead to large difference gain values especially when $n_i$ is larger, which indicates exploded certified bounds at initialization.
In contrast, our initialization yields a constant difference gain of 1 regardless of $n_i$.

While \citet{he2015delving} proposed Kaiming initialization to stabilize the variance of each layer in standard DNN training, compared to Xavier initialization, it has even larger difference gain values and thus it tends to worsen the tightness of certified bounds here.
For CNN-7 on CIFAR-10 using 160 total training epochs, if we make the Vanilla IBP baseline use Kaiming initialization, the verified error is $(68.07\pm 0.30)\%$, which is worse than the baseline result using Xavier initialization, i.e., $ (67.01\pm 0.29)\% $. The empirical result aligns with our theoretical insight since Kaiming initialization has larger difference gain values.

\subsection{Illustration of IBP Relaxations for Different Neuron States}

\begin{figure}[ht]
\centering
\includegraphics[width=.8\textwidth]{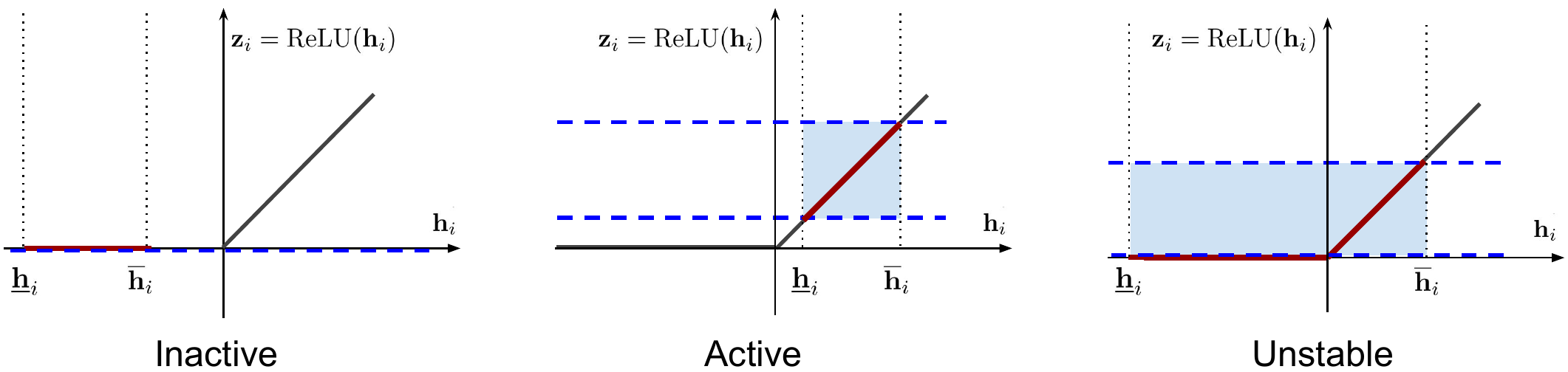}
\caption{Three activation states of ReLU neurons determined by pre-activation lower and upper bounds and their corresponding IBP relaxations. The relaxed areas are shown in light blue.
}
\label{fig:relu_neuron_states}
\end{figure}

In Figure~\ref{fig:relu_neuron_states}, we illustrate IBP relaxations for ReLU neurons with the three different states respectively.
Inactive neurons have no relaxation error compared with the other two kinds of neurons, and thus IBP training tends to prefer inactive neurons more to tighten certified bounds, compared to the other two ReLU neuron states. This leads to an imbalance in ReLU neuron states for vanilla IBP on models without BN. In this paper, we identify the benefit of fully adding BN layers to mitigate the imbalance, because BN normalizes pre-activation values. We also add a regularization to further encourage ReLU balance.

\subsection{IBP Initialization for Non-feedforward Networks}
\label{apd:init_resnet}

Our analysis in Section~\ref{sec:exploded_bounds_at_init} is based on feedforward networks but it can also be easily extended to other architectures.
On the weight initialization for standard DNN training, \citet{NEURIPS2018_d81f9c1b,NEURIPS2019_e520f70a} extended the weight initialization to ResNet, which aimed to keep the variance stable. 
In IBP initialization, we want to make $\E(\Delta_i)$ stable instead, and we give an example on ResNet.
We consider a residual connection $\tilde{\rvh}_i= \rvh_i+\rvh_{i-1}$, and we want to make its tightness $ \ol{\rvh}_i+\ol{\rvh}_{i-1} - (\ul{\rvh}_i+\ul{\rvh}_{i-1}) $ stable, which equals to $\Delta_i + \Delta_{i-1}$. 
Our IBP initialization in Section~\ref{sec:init} makes $ \E(\Delta_i)\approx\E(\Delta_{i-1})$, and thereby $ \E(\Delta_i+\Delta_{i-1})\approx 2\E(\Delta_{i-1})$. Here we get an additional growth factor of 2, when propagating bounds from layer $i-1$ to layer $i$. This factor is a constant and does not depend on the fan-in number $n_i$. 
We can further remove this factor, we can divide the weight after each residual connection by 2 (this is equivalent to dividing $ \tilde{\rvh}_i $ by 2 when it is used by subsequent layers).

\subsection{Effect of Batch Normalization on IBP Bound Tightness}
\label{ap:BN_IBP}

Our analysis in Section~\ref{sec:exploded_bounds_at_init} does not consider BN.
In this section, we analyze the tightness of certified bounds when BN presents. As mentioned in Section~\ref{sec:bn}, we use mean and variance estimation computed from clean data in BN (which is also the standard way). 
For the output bounds $\ul{\rvh}_i$ and $\ol{\rvh}_i$, we use $\underline{\rvh}_i'$ and $\overline{\rvh}_i'$ to denote the output bounds after BN.
We have $\underline{\rvh}_i' = a_i \frac{\underline{\rvh}_i - \mu(\boldsymbol{\rvh}_i)}{\sigma(\boldsymbol{\rvh}_i)} + b_i $, 
where $\mu(\boldsymbol{\rvh}_i)$ and $\sigma(\boldsymbol{\rvh}_i) $ stand for the estimated the mean and standard deviation respectively from clean output $\rvh_i $, and $a_i$ and $b_i$ are the weight and bias of BN. 
Similarly we can get $\overline{\boldsymbol{\rvh}}_i'$. Therefore, to conduct a analysis similar to Sec. 3.2.1 for BN, we first need to estimate $\mu(\boldsymbol{\rvh}_i)$ and $\sigma(\boldsymbol{\rvh}_i)$, and then we can estimate $\overline{\boldsymbol{\rvh}}_i', \underline{\boldsymbol{\rvh}}_i'$. Finally, we have $\Delta_i' = \overline{\boldsymbol{\rvh}}_i' - \underline{\boldsymbol{\rvh}}_i'$ to denote the bound tightness after BN.

At initialization, we assume elements in $\rvh_i$ are independently initialized following a zero-mean Gaussian distribution, and $\Delta_i'$ can be computed from the variance of the Gaussian distribution.
However, after a single step of training, elements in $\rvh_i$ are no longer independent, and the mean and variance in BN are difficult to calculate explicitly.
But we can empirically estimate them. Although when $\sigma(\boldsymbol{\rvh}_i) < 1$ (which is true if we use IBP initialization to tighten certified bounds), $\Delta_i’$ will get larger than $\Delta_i$, i.e., bounds become looser after they are propagated through BN, we can show empirically that IBP initialization is still able to tighten the bounds in this situation.

In Table~\ref{tab:E}, we compare $\log(\hat{\mathbb{E}}(\Delta_m)/\hat{\mathbb{E}}(\Delta_0))$ of CNN-7 model with full BN on CIFAR-10 during the early epochs, where $m$ is the last layer of the model, with and without IBP initialization respectively. A smaller value indicates that the bounds are tighter. And we can see that the model with IBP initialization has smaller $\hat{\mathbb{E}}(\Delta_m)/\hat{\mathbb{E}}(\Delta_0)$ along these epochs and thus has tighter bounds.

\begin{table}[h]
    \centering
    \begin{tabular}{c|c|c|c|c}
    \toprule[1pt]
         IBP Initialization & Epoch 1 & Epoch 2 & Epoch 3 & Epoch 4  \\
    \midrule
         No& 16.29& 15.21 & 13.08 & 11.90\\
         Yes& 11.56 & 12.42& 11.97& 11.24\\
    \bottomrule[1pt]
    \end{tabular}
    \caption{$\log(\hat{\mathbb{E}}(\Delta_m)/\hat{\mathbb{E}}(\Delta_0))$ at the first 5 epochs of CNN-7  with full BN on CIFAR-10, with and without IBP initialization respectively, which reflects the tightness of certified bounds along the training}
    \label{tab:E}
\end{table}

\section{Additional Experiments}
\label{ap:exp}

\subsection{Computational Cost for All Datasets and Models}
\label{ap:time}

In addition to the time cost comparison on CNN-7 on CIFAR shown in Section~\ref{sec:time}, we report computation cost results for all the datasets and models in Table~\ref{tab:time_full}.
Under same training schedules, results show that our proposed method has a small overhead over vanilla IBP, and the cost is still lower than that of CROWN-IBP.
Meanwhile, our method is able to achieve lower verified errors compared to the two baselines (Table~\ref{tab:main} and Table~\ref{tab:tiny_imagenet}).
More importantly, we are able to use much shorter training schedules to achieve SOTA results compared to previous literature, which enables faster certified robust training.

\begin{table*}[ht]
    \centering
    \caption{
    Comparison of estimated time cost (seconds) on all the datasets and models. We report the per-epoch time during training phases with different $\eps$ ranges, and we report the total time when the 70-epoch schedule is used for MNIST, the 160-epoch schedule for CIFAR-10, and the 80-epoch schedule for TinyImageNet respectively. ``-'' in the table means that there is no $\eps=0$ warmup stage for MNIST following \citet{zhang2019towards}. Note that on each dataset, for phases of same or different methods that are supposed to be equivalent in algorithm implementation, we make them share the same time estimation result respectively. 
   }
   \adjustbox{max width=.8\textwidth}{
   \begin{threeparttable}
    \begin{tabular}{c|cc|cccc}
    \toprule
    
    \multirow{2}{*}{Dataset} & \multirow{2}{*}{Model} &  \multirow{2}{*}{Method} & \multicolumn{3}{c}{Per-epoch for $\eps$} & \multirow{2}{*}{Total} \\
    & & &  $0$ & $(0,\epst)$ & $\epst$ \\
    \hline
    
    \multirow{9}{*}{MNIST} & \multirow{3}{*}{CNN-7} & Vanilla IBP & - & 27.9 & 27.9 & 1955.1  \\
    & & CROWN-IBP & - & 49.6 & 27.9 & 2387.5 \\
    & & Ours& - & 37.0 & 27.9 & 2135.8 \\
    \cline{2-7}
    
    & \multirow{3}{*}{Wide-ResNet} & Vanilla IBP & - & 81.0 & 81.0 & 5668.3  \\
    & & CROWN-IBP & - & 142.1 & 81.0 & 6890.2 \\
    & & Ours& - & 99.0 & 81.0 & 6029.3 \\
    \cline{2-7}
    
    & \multirow{3}{*}{ResNeXt} & Vanilla IBP & - & 73.2 & 73.2 & 5127.2  \\
    & & CROWN-IBP & - &  147.7 & 73.2 & 6616.9 \\
    & & Ours& - & 104.4 & 73.2  & 5750.7\\
    \hline
    
    \multirow{9}{*}{CIFAR-10} & \multirow{3}{*}{CNN-7} & Vanilla IBP &  30.0 & 54.8 & 54.8 & 8747.9 \\
    & & CROWN-IBP & 30.0 & 78.5 & 54.8 & 10641.3 \\
    & & Ours& 64.0 & 64.0 & 54.8 & 9512.3 \\
    \cline{2-7}
    
    & \multirow{3}{*}{Wide-ResNet} & Vanilla IBP & 43.7 & 114.7 & 114.7 & 18358.4  \\
    & & CROWN-IBP &43.7 & 170.7 & 114.7 & 22764.9 \\
    & & Ours& 134.7 & 134.7 & 114.7 & 19976.0 \\
    \cline{2-7}
    
    & \multirow{3}{*}{ResNeXt} & Vanilla IBP & 38.7 & 102.7 & 102.7 & 16432.0  \\
    & & CROWN-IBP & 38.7 & 183.3 & 102.7 & 22813.6 \\
    & & Ours& 129.6 & 129.6 & 102.7 & 18611.7\\
    \hline
    
    \multirow{9}{*}{TinyImageNet} & \multirow{3}{*}{CNN-7} & Vanilla IBP & 282.2 & 431.4 & 431.4 & 34362.0\\
    & & CROWN-IBP & 282.2 & 663.8 & 431.4 & 36686.5 \\
    & & Ours&  500.4 & 500.4 & 431.4 & 35270.3 \\
    \cline{2-7}
    
    & \multirow{3}{*}{Wide-ResNet} & Vanilla IBP & 270.2 & 399.8 & 399.8 & 31861.6  \\
    & & CROWN-IBP & 270.2 & 592.1 & 399.8 & 33789.3\\
    & & Ours& 464.6 & 464.6 & 399.8 & 32703.0\\
    \cline{2-7}
    
    & \multirow{3}{*}{ResNeXt} & Vanilla IBP & 197.2 & 430.5 & 430.5 & 34206.7  \\
    & & CROWN-IBP &197.2 & 883.1 & 430.5 & 38735.1 \\
    & & Ours& 626.3 & 626.3 & 430.5 & 36595.8\\
    \bottomrule
    \end{tabular}
    \end{threeparttable}}
    \label{tab:time_full}
\end{table*}

\subsection{Additional Ablation Study}
\label{ap:additional_ablation}

\begin{table*}[ht]
\centering
\caption{
Additional ablation study results on BN where we consider whether centralization and unitization in BN present respectively.
The results are from CNN-7 on CIFAR-10 ($\epst=8/255$) using the training schedule with 160 epochs in total. We compare the proportion of active ReLU neurons and inactive ReLU neurons respectively, and also the errors.
}
\adjustbox{max width=\textwidth}{
\begin{tabular}{cccccc}
\toprule
Centralization & Unitization & Active ReLU (\%) & Inactive ReLU (\%) & Standard error (\%) & Verified error (\%)\\
\midrule

$\times$ & $\times$ &  7.37$\pm$0.25 &	90.57$\pm$0.30 & 57.36$\pm$0.45 &	69.91$\pm$0.31 \\

$\checkmark$ & $\times$ & 13.48$\pm$0.22	& 84.73$\pm$0.26 &	55.36$\pm$0.17 &	68.07$\pm$0.02\\

$\times$ & $\checkmark$ & 16.94$\pm$0.79	& 80.40$\pm$0.75 & 54.41$\pm$0.49 & 	67.78$\pm$0.46\\

$\checkmark $ & $\checkmark$ & 21.30$\pm$0.39 & 75.90$\pm$0.40 & 51.72$\pm$0.40 & 65.58$\pm$0.30\\

\bottomrule
\end{tabular}}
\label{tab:ablation_bn}
\end{table*}

In this section, we present additional ablation study results on BN, where we split the centralization (the shifting operation using the mean) and the unitization (the scaling operation using the variance) to investigate whether both of them contribute to the improvement by BN.
We run this experiment for CNN-7 on CIFAR-10 ($\epst=8/255$) using the training schedule with 160 epochs in total, and we show the results in Table~\ref{tab:ablation_bn}.

From our ablation results, we can observe that both centralization and unitization in BN contribute to the improvement. We conclude the benefit as follows. First, BN has inherent benefits for standard DNN training~\citep{ioffe2015batch,van2017l2,santurkar2018does}. In addition, BN benefits IBP also because it has an effect on balancing ReLU neuron states, as our results show that when a model is trained with BN, the number of active ReLU neurons is noticeably better than the cases without BN. We found that actually both mean centralization and unitization help to balance active and inactive ReLU neurons. It is easy to understand that centralization helps balancing as it can center the bounds around zero. For unitization, we conjecture that it helps the optimization for DNN (from the acceleration or smoothing the loss landscape perspective for standard DNN training), and this may allow the model to have a less tendency to reduce the robust loss by trivially making most neurons inactive.

\subsection{Other Perturbation Radii}
\label{ap:multi_eps}

In Table~\ref{tab:multi_eps}, we present results using perturbation radii other than those used in our main experiments.
Here we consider $\epst\in\{0.1,0.3\}$ for MNIST, and $\epst\in\{\frac{2}{255},\frac{16}{255}\}$ for CIFAR-10.
In particular, on MNIST models are trained with target perturbation radii $\eps_{\text{train}}$  larger than used for testing $\epst$ to mitigate overfitting -- we use $\eps_{\text{train}}=0.2$ when $\epst=0.1$ and $\eps_{\text{train}}=0.4$ when $\epst=0.3$ following \citet{zhang2019towards}.
We use CNN-7 in this experiment.
Results show that improvements over Vanilla IBP and CROWN-IBP are consistent as in Table~\ref{tab:main}.
Note that CIFAR-10 with very small $\eps=\frac{2}{255}$ is a special case where using pure linear relaxation bounds~\citep{wong2018provable,zhang2019towards} for training yields even lower errors than IBP~\citep{gowal2018effectiveness} and standard CROWN-IBP which anneals to IBP training after warmup. On this setting, an alternative version of CROWN-IBP that does not anneal to IBP training can achieve lower verified error 43.61\% without loss fusion (60.44\% if loss fusion is enabled). However, using pure linear relaxation bounds for certified training is more costly and usually has worse results on other settings~\citep{jovanovic2021certified}. Thus for all the other settings in \citet{zhang2019towards}, CROWN-IBP still have to anneal to IBP training, as the version we adopt in our main experiments. 
Overall, the experimental results demonstrate that our proposed method is effective on settings with different perturbation radii, compared to vanilla IBP and CROWN-IBP.

\begin{table}[ht]
  \centering
  \caption{
  The standard errors (\%) and verified errors (\%) of a CNN-7 model trained with different methods on other perturbation radii not included in the main results.
}
\label{tab:multi_eps}
  \adjustbox{max width=\textwidth}{
  \begin{threeparttable}	
  \begin{tabular}{cccccccccc}
    \toprule[1pt]
    \multirow{2}{*}{Dataset} & \multirow{2}{*}{Warmup} &  \multirow{2}{*}{$\epst$} & \multirow{2}{*}{$\eps_{\text{train}}$} & \multicolumn{2}{c}{Vanilla IBP} & \multicolumn{2}{c}{CROWN-IBP} & \multicolumn{2}{c}{Ours}\\
    & & & & Standard & Verified & Standard & Verified & Standard & Verified\\
    \hline
    \multirow{2}{*}{MNIST} & \multirow{2}{*}{$0\!+\!20$}& 0.1 & 0.2 & 1.12 & 2.17 & 1.07 & 2.17 & 1.16 & \textbf{2.05} \\
    & &  0.3 & 0.4 & 2.74 & 7.61 & 2.88 & 7.55 & 2.33 & \textbf{6.90}\\
    \hline
    \multirow{2}{*}{CIFAR-10} & \multirow{2}{*}{$1\!+\!80$} &  \multicolumn{2}{c}{2/255} & 33.65 & 48.75 & 34.09 & 48.28 & 33.16 & \textbf{47.15} \\
    & & \multicolumn{2}{c}{16/255} & 64.52 & 76.36 & 71.75 & 79.43 & 63.35 & \textbf{75.52}\\
    \bottomrule[1pt]
  \end{tabular}
    \end{threeparttable}}
  \end{table}

\subsection{Sensitivity on the  $\lambda_0$ Hyperparameter}
To test the sensitivity of the training performance on the choice of $\lambda_0$, we run an experiment on CNN-7 for CIFAR-10 ($\epst=8/255$) using the $ 160 $-epoch schedule. We consider $\lambda_0\in\{ 0.1,0.2,0.5,1.0,2.0\} $, and we run 5 repeated experiments for each setting to report the mean and standard deviation. We show the results in Table~\ref{tab:lambda}.

We find that $\lambda_0=0.5$ or $\lambda_0=1.0$ both yield good results on this setting. Actually, for all the results of “ours” in Table~\ref{tab:main} (MNIST and CIFAR-10) in the paper, we always use $\lambda_0=0.5$ for all settings, and we do not tune $\lambda_0$ for each setting individually. This suggests that potential users do not need to search for $\lambda_0$ in each training. 
Similarly, on TinyImageNet, good results can be achieved by using $\lambda_0=0.1$ for all the settings. The $\lambda_0$ for TinyImageNet is smaller, and this can be explained by smaller $\epst$ for TinyImageNet ($1/255$) compared to 0.4 for MNIST and $8/255$ for CIFAR-10.
Thus, the results suggest that our approach is not very sensitive to choice of $\lambda_0$, and a reasonable default value can work well for many settings (e.g., under many different training schedules or models).

\begin{table*}[ht]
\centering
\caption{
Results of the sensitivity test for the $\lambda_0$ hyperparamter, on CNN-7 for CIFAR-10 ($\epst=8/255$) using the $ 160 $-epoch schedule.
}
\begin{tabular}{c|ccccc}
\toprule
$\lambda_0$ & 0.1 & 0.2 & 0.5 & 1.0 & 2.0\\
\midrule
Standard error (\%) &  53.03 $\pm$ 0.56 & 53.08 ± 0.62 & 51.72±0.40 & \textbf{50.98$\pm$0.33} & 53.80$\pm$ 0.37\\
Verified error (\%) & 66.44 $\pm$ 0.24 & 66.54 $\pm$ 0.48 & \textbf{65.58$\pm$0.32} & \textbf{65.42$\pm$0.22} &  66.91$\pm$0.26\\
\bottomrule
\end{tabular}

\label{tab:lambda}
\end{table*}

\subsection{Applying the Proposed Method to CROWN-IBP}

We have tried applying our method to CROWN-IBP~\citep{zhang2019towards} besides IBP. 
On CNN-7 for CIFAR-10 ($\epst=8/255$) with the $160$-epoch training schedule, we observe that adding BN improves the performance of CROWN-IBP (verified error $68.02\%\rightarrow66.93\%$ if loss fusion is disabled; $76.11\%\rightarrow 68.8\%$ if loss fusion is enabled).
However, further adding IBP initialization or the warmup regularizers does not significantly change the performance. 
For the possible reasons of this result, we analyze that: 1) CROWN-IBP already has tighter bounds by a linear relaxation based bound propagation; 2) CROWN-IBP has tight relaxation for both inactive and active ReLU neurons, compared to IBP which has tight relaxation only for inactive neurons but not active ones, so the imbalanced ReLU issue is less significant for CROWN-IBP (For the setting in Figure~\ref{fig:unbalanced_relu}, we empirically find that even if we do not use warmup regularization, CROWN-IBP already has around 19\% active neurons, even more than our improved IBP).
Thus, it is reasonable that our proposed method focusing on improving bound tightness and ReLU neuron balance may be less effective for CROWN-IBP. 

Instead, there may be other factors that limit the performance of linear relaxation based certified robust training.
So far tighter linear or convex relaxation bounds (e.g., \citet{zhang2018efficient} or \citet{wong2018provable}) usually cannot outperform pure IBP using looser interval bounds. 
While \citet{zhang2019towards} used linear relaxation bounds for certified training and outperformed pure IBP, their method still needs to gradually anneal to pure IBP bounds in the end of training.
There are some recent works that studied the reasons behind this phenomenon.
\citet{jovanovic2021certified} identified two properties of convex relaxations, continuity and sensitivity, that may impact training dynamics.s
\citet{lee2021loss} identified a factor about the smoothness of loss landscape, and they proposed to use tighter relaxation via optimizing the bounds, which may lead to more favorable loss landscapes. 
In terms of improving the verified errors after training, \citet{jovanovic2021certified} only have preliminary results on a small network for MNIST since their new relaxations require solving convex/linear programs; and we outperform \citet{lee2021loss} (15.42\% for MNIST $\epst=0.4$; 69.70\% for CIFAR-10 $\epst=8/255$) with a notable margin, while we use much shorter training schedules.

\subsection{Comparison with Randomized Smoothing}

In this section, we empirically compare the performance of our method with randomized smoothing methods.
As we have mentioned in Sec.~\ref{sec:related_work}, 
randomized smoothing is mostly suitable for $\ell_2$-norm certified defense, and it is fundamentally limited for $\ell_\infty$ norm robustness.
But there are still existing works such as \citet{salman2019provably} that use norm inequalities to convert $\ell_2$ norm robustness certificates to an $\ell_\infty$ norm one.
For $\ell_\infty$-norm perturbation radius $\eps=\frac{8}{255}$ on CIFAR-10 where each input image has $ 3\times 32\times 32 $ dimensions, we can convert it to $\ell_2$-norm radius $ \eps_2 = \frac{8}{255} \times \sqrt{3\times 32\times 32} = 1.73884 $ used by randomized smoothing such that the certified accuracy under this $\ell_2$ perturbation provides a lower bound for $\ell_\infty$ certified robustness under radius $\eps=\frac{8}{255}$. 
In an earlier work~\citep{li2018second}, their certified accuracy under this perturbation size is 0, according to their Figure 1; and in a more recent work~\citep{salman2019provably}, according to results in their Table 7, their best certified accuracy is 23\% for radius 1.75, and the certified accuracy is 26\% for radius 1.5, so their certified error is at least 74\% for $\ell_\infty \epsilon=8/255$. 
Therefore, the certified error we achieve in our paper is much lower (65.58±0.32\%), compared to randomized smoothing by converting $\ell_2$ certified radius.

\subsection{ReLU Imbalance with Shorter Warmup Length}
\label{ap:relu_balance_short}

\begin{figure}[ht]
	\centering
	\includegraphics[width=0.5\textwidth]{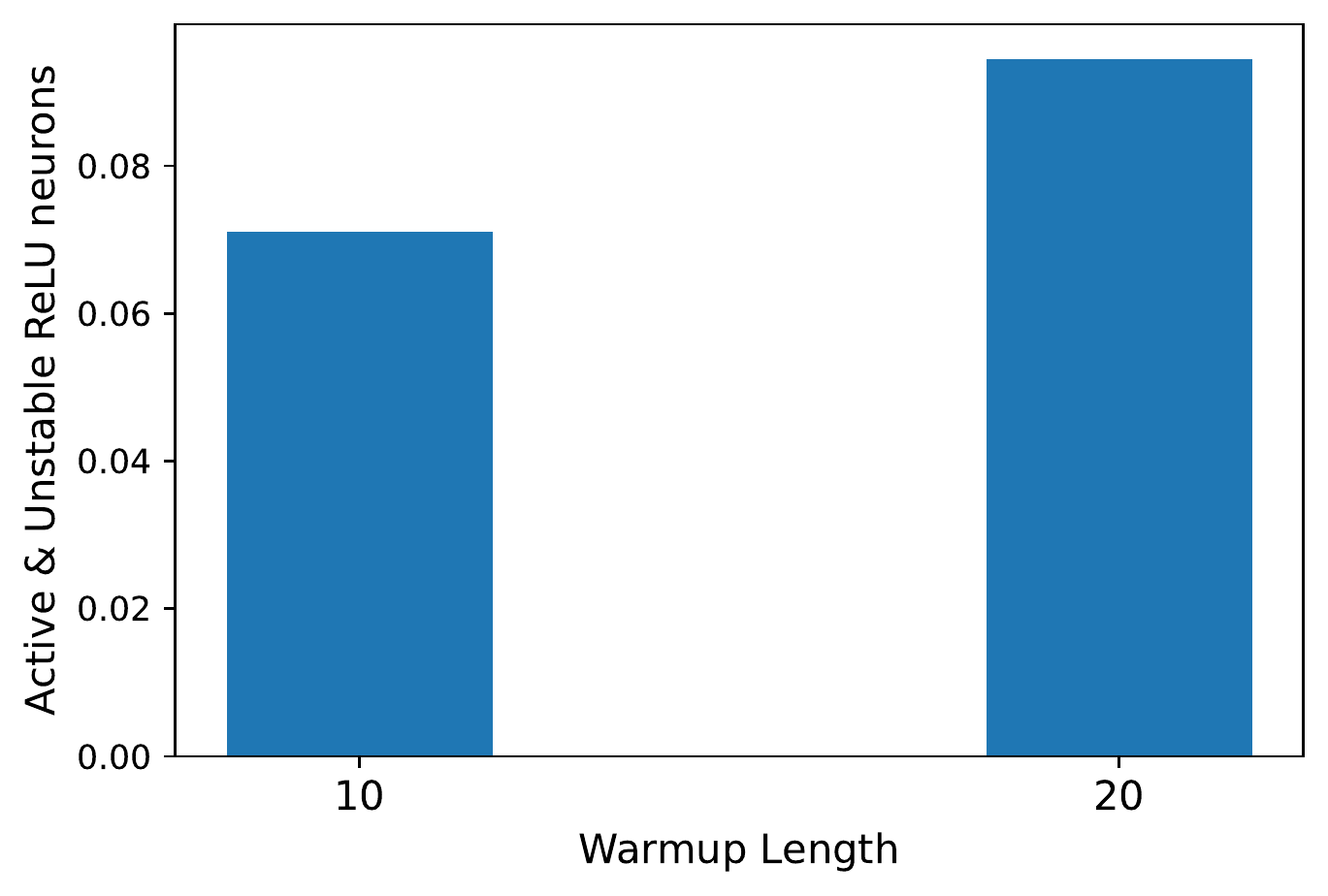}
	\caption{Ratio of active and unstable neurons in CNN-7 trained with Vanilla IBP using different warmup lengths respectively.
	}
	\label{fig:unbalenced_relu_shorter}
\end{figure}
 
In Figure~\ref{fig:unbalanced_relu}, we show two 7-layer CNN models with different warmup length respectively, and the model tends to have more inactive neurons and thus more severe imbalance in ReLU neuron states for shorter warmup length, as previously mentioned in Section~\ref{sec:imbalance_relu}.

\subsection{Using IBP Initialization when Bound Explosion is More Severe}
\label{ap:resnext_explosion}

\begin{figure}[ht]
\centering
\includegraphics[width=0.5\textwidth]{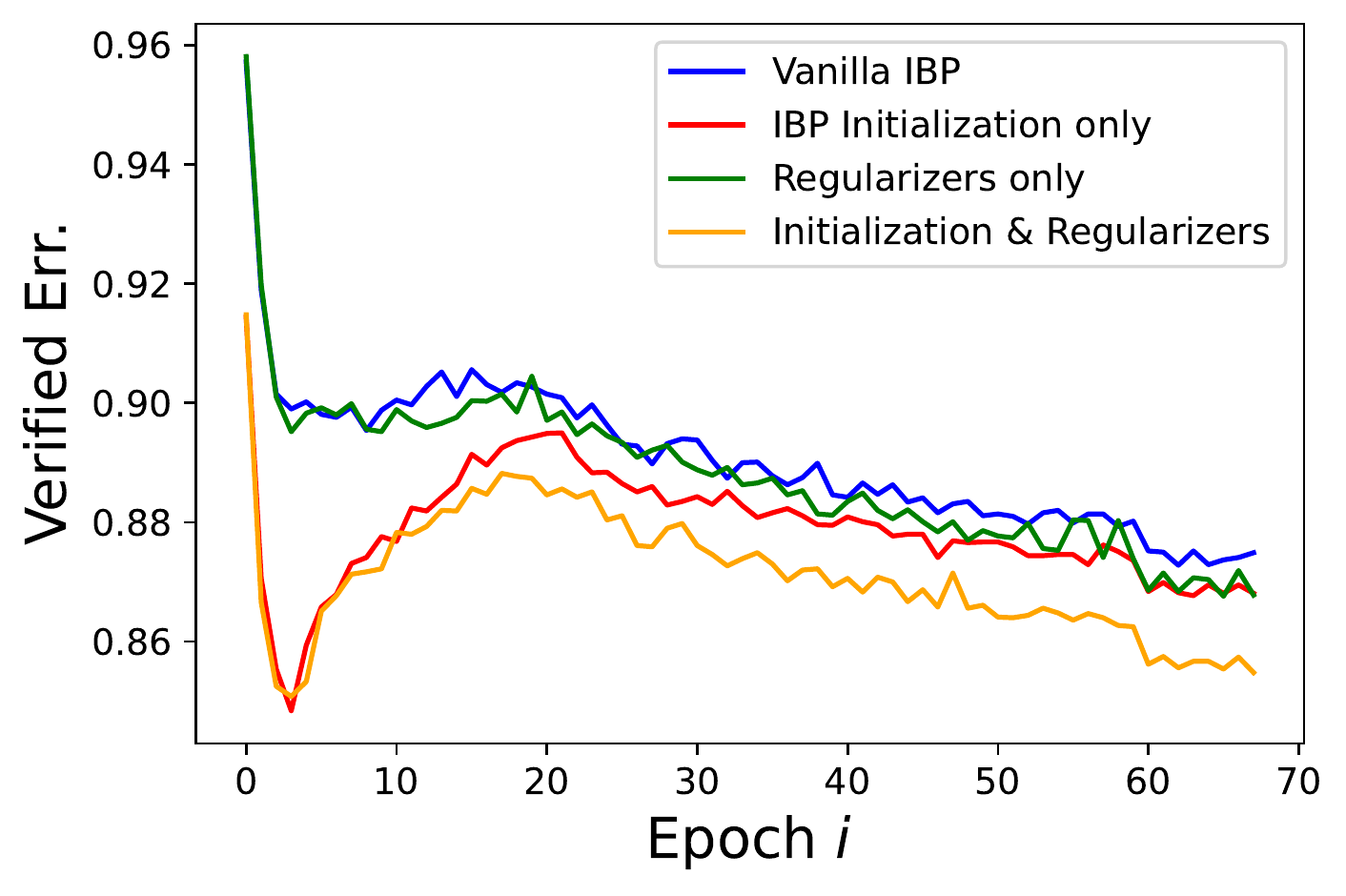}
\caption{Curve of training verified error of a ResNeXt model on TinyImageNet. Note that the verified errors can increase during the warmup as $\eps$ increases.}
  \label{fig:l_resnext}
\end{figure}

In Figure~\ref{fig:l_resnext}, we show that for a ResNeXt on TinyImageNet, where the explosion of certified bounds is more severe if the network is initialized with standard weight initialization, using our proposed initialization is helpful for reaching lower verified errors especially at early epochs.

\section{Experimental Details}
\label{ap:imp}

\paragraph{Implementation} Our implementation is based on the \texttt{auto\_LiRPA} library~\citep{xu2020automatic}\footnote{\url{https://github.com/KaidiXu/auto_LiRPA}} which supports robustness verification and training on general computational graphs.
Baselines including Vanilla IBP and CROWN-IBP with loss fusion are inherently supported by the library.
We add to implement our IBP initialization and warmup with regularizers for fast certified robust training. 

\paragraph{Datasets} For MNIST and CIFAR-10, we load the datasets using \texttt{torchvision.datasets}\footnote{\url{https://pytorch.org/vision/0.8/datasets.html}} and use the original data splits.
On CIFAR-10, we use random horizontal flips and random cropping for data augmentation, and also normalize input images, following \citet{zhang2019towards,xu2020automatic}.
For TinyImageNet, we download the dataset from Stanford CS231n course website\footnote{\url{http://cs231n.stanford.edu/TinyImageNet-200.zip}}.
Similar to CIFAR-10, we also use data augmentation and normalize input images for TinyImageNet.
Unlike \citet{xu2020automatic} which cropped the $64\times 64$ original images into $56\times 56$ and used a central $56\times 56$ cropping for test images, we pad the cropped training images back to $64\times 64$ so that we do not need to crop test images.
We use the validation set for testing since test images are unlabelled, following \citet{xu2020automatic}.

\paragraph{Models}
We use three model architectures in the experiments: a 7-layer feedforward convolutional network (CNN-7), Wide-ResNet~\citep{zagoruyko2016wide} and ResNeXt~\citep{xie2017aggregated}.
All the models have a hidden fully-connected layer with 512 neurons prior to the classification layer.
For CNN-7, there are five convolutional layers with $64,64,128,128,128$ filters respectively.
For Wide-ResNet, there are 3 wide basic blocks, with a widen factor of 8 for MNIST and CIFAR-10 and 10 for TinyImageNet.
For ResNeXt, we use $1,1,1$ blocks for MNIST and CIFAR-10, and $2,2,2$ blocks for TinyImageNet; the cardinality is set to 2, and the bottleneck width is set to 32 for MNIST and CIFAR-10 and 8 for TinyImageNet.
For all the models, ReLU is used as the activation.
These models were similarly adopted in \citet{xu2020automatic}.
But we fully add BNs after each convolutional layer and fully-connected layer, while some of these BNs were missed in \citet{xu2020automatic}.
For example, the CNN-7 model in \citet{xu2020automatic} had BN for convolutional layers but not the fully-connected layer.
Besides, we remove the average pooling layer in Wide-ResNet as we find it harms the performance of all the considered training methods, and this modification makes the Wide-ResNet align better with the CNN-7 model, which does not have average pooling either and achieves best results compared to other models (Table~\ref{tab:main} and Table~\ref{tab:tiny_imagenet}).

\paragraph{Training} During certified training, models are trained with Adam~\citep{kingma2014adam} optimizer with an initial learning rate of $5\times 10^{-4}$, and there are two milestones where the learning rate decays by 0.2. 
We determine the milestones for learning rate decay according to the training schedule and the total number of epochs, as shown in Table~\ref{tab:lr_decay}.
Gradient clipping threshold is set to 10.0.
We train the models using a batch size of 256 on MNIST, and 128 on CIFAR-10 and TinyImageNet.
The tolerance value $\tau$ in our warmup regularization is fixed to 0.5.
For Vanilla IBP and IBP with our initialization and regularizers, we train the models on a single NVIDIA GeForce GTX 1080 Ti or NVIDIA GeForce RTX 2080 Ti GPU for each setting.
For CROWN-IBP, we train the models on two GPUs for efficiency, while in time estimation we still use one single GPU for fair comparison.
The number of training and evaluation runs is 1 for each experiment result respectively.
In the evaluation, the major metric is \emph{verified error}, which stands for the rate of test examples such that the model cannot certifiably make correct predictions given the $\ell_\infty$ perturbation radius. For reference, we also report \emph{standard error}, which is the standard error rate where no perturbation is considered.

\begin{table}[ht]
  \centering
    \caption{Milestones for learning rate decay when different total number of epochs are used. ``Decay-1'' and ``Decay-2'' denote the two milestones respectively when the learning rate decays by a factor of 0.2. 
   }
  \footnotesize
  \adjustbox{max width=.7\textwidth}{
  \begin{tabular}{cccc}
    \toprule[1pt]
    
    Dataset & Total epochs & Decay-1 & Decay-2\\
    \midrule
    
    \multirow{2}{*}{MNIST} & 50 & 40 & 45\\
    & 70 & 50 & 60\\
    \hline
    
    \multirow{2}{*}{CIFAR-10} & 70 & 50 & 60\\
    & 160 & 120 & 140\\
    
    \hline
    
    {TinyImageNet} & 80 & 60 & 70\\   
    
    \bottomrule[1pt]
  \end{tabular}
    }
  \label{tab:lr_decay}
\end{table}

\paragraph{Warmup scheduling} During the warmup stage, after training with $\eps=0$ for a number of epochs, the perturbation radius $\eps$ is gradually increased from 0 until the target perturbation radius $\epst$, during the $0\!<\!\eps\!<\!\epst$ phase.
Specifically, during the first 25\% epochs of the $\eps$ increasing stage, $\eps$ is increased exponentially, and after that $\eps$ is increased linearly.
In this way, $\eps$ remains relatively small and increases relatively slowly during the beginning, to stabilize training.
We use the \texttt{SmooothedScheduler} in the \texttt{auto\_LiRPA} as the scheduler for $\eps$ similarly adopted by \citet{xu2020automatic}.
On CIFAR-10, unlike some prior works which made the perturbation radii used for training $1.1$ times of those for testing respectively~\citep{gowal2018effectiveness,zhang2019towards}, we find this setting makes little improvement over using same perturbation radii for both training and testing in our experiments as also mentioned in \citet{lee2021loss}, and thus we directly adopt the later setting for simplicity.

\section{Mathematical Proofs}

\label{ap:proofs}

\subsection{Proof of \eqref{eq:e_grow}}
\label{apd:proof_delta}

In this section, we provide a proof for \eqref{eq:e_grow}:
\begin{equation}
        \E(\delta_{i}) = \E(\text{ReLU}(\overline{\rvh}_{i})) - \text{ReLU}(\underline{\rvh}_{i}))=\frac{1}{2} \E(\Delta_{i}),
\end{equation}
where $\Delta_i=\overline{\rvh}_i-\underline{\rvh}_i$, and $\delta_i=\overline{\rvz}_i-\underline{\rvz}_i$.

\begin{proof}


We first have
    \begin{equation}
        \begin{aligned}
            \E(\delta_i)
            =& \E(\text{ReLU}(\overline{\rvh}_i)\!-\! \text{ReLU}(\underline{\rvh}_i))\\
            =& \E(\text{ReLU}(\rvc_i\!+\! \frac{\Delta_i}{2})\!-\! \text{ReLU}(\rvc_i\!-\! \frac{\Delta_i}{2}))\\
            =& \E(\text{ReLU}(\rvc_i\!+\! \frac{\Delta_i}{2}))\!-\! \E(\text{ReLU}(\rvc_i\!-\! \frac{\Delta_i}{2})).
        \end{aligned}
    \end{equation}
Note that $\rvc_i = \frac{1}{2}\rmW_i(\underline{\rvz}_i + \overline{\rvz}_i)$ and $\Delta_i = |\rmW_i|\delta_i$, and thus $p(-\rvc_i~|~|\rmW_i|) = p(\rvc_i~|~|\rmW_i|)$ and $p(-\rvc_i|\Delta_i) = p(\rvc_i|\Delta_i)$, where we use $p(\cdot)$ to denote the probability density function (PDF).
Thereby,
\begin{equation}
    \begin{aligned}
        &\E(\text{ReLU}(\rvc_i + \frac{\Delta_i}{2})) = \int_{0}^\infty \int_{-\frac{\Delta_i}{2}}^\infty (\rvc_i + \frac{\Delta_i}{2}) p(\rvc_i|\Delta_i) p(\Delta_i) d \rvc_i d\Delta_i,\\
        &\E(\text{ReLU}(\rvc_i - \frac{\Delta_i}{2})) = \int_{0}^\infty \int_{\frac{\Delta_i}{2}}^\infty (\rvc_i - \frac{\Delta_i}{2}) p(\rvc_i|\Delta_i) p(\Delta_i) d \rvc_i d\Delta_i.
    \end{aligned}
\end{equation}
And thus
\begin{equation}
    \begin{aligned}
        &\E(\text{ReLU}(\rvc_i\!+\! \frac{\Delta_i}{2}))\!-\! \E(\text{ReLU}(\rvc_i\!-\! \frac{\Delta_i}{2}))\\
        =& \int_{0}^\infty (\int_{\frac{\Delta_i}{2}}^\infty \Delta_i + \int_{-\frac{\Delta_i}{2}}^{\frac{\Delta_i}{2}} (\rvc_i + \frac{\Delta_i}{2})) p(\rvc_i|\Delta_i) p(\Delta_i) d \rvc_i d\Delta_i\\
        =& \int_{0}^\infty \int_{-\infty}^{\infty}\frac{\Delta_i}{2}p(\rvc_i|\Delta_i) p(\Delta_i) d \rvc_i d\Delta_i\\
        =& \frac{1}{2} \E(\Delta_i).
    \end{aligned}
\end{equation}
\end{proof}

\subsection{Proof on the Bounds of $ \Var(\underline{\rvh}_i)$ and $\Var(\overline{\rvh}_i)$  }
\label{apd:bound_variance}

In this section, we show that $ \Var(\underline{\rvh}_i)$ and $\Var(\overline{\rvh}_i)$ will not explode or vanish at initialization, so that the magnitude of forward signals will not vanish or explode when we use IBP initialization which focuses on stabilizing the tightness of certified bounds.

We can derive that
\begin{align*}
       \Var(\overline{\rvh}_{i}) &= \Var(\rmW_{i,+} \overline{\rvz}_{i-1} + \rmW_{i,-}\underline{\rvz}_{i-1})\\
       &= \Var([\rmW_{i,+} \overline{\rvz}_{i-1} + \rmW_{i,-}\underline{\rvz}_{i-1}]_j) \enskip (0\leq j\leq r_i)\\
       &= \Var\bigg(\sum\nolimits_{k = 1}^{n_i}([\rmW_i]_{j, k} [\overline{\rvz}_{i-1}]_k \cdot \mathbb{I}([\rmW_i]_{j, k} > 0))
       \\
       &\qquad\enskip+ \sum\nolimits_{k = 1}^{n_i}([\rmW_i]_{j, k} [\underline{\rvz}_{i-1}]_k \cdot \mathbb{I}([\rmW_i]_{j, k} \leq 0))\bigg).
 \end{align*}
Since $\rmW_i$ is initialized with mean $0$, the numbers of negative elements and positive elements are approximately equal, and thus
 \begin{align*}
 	\Var(\overline{\rvh}_{i}) &\approx \frac{n_i}{2} \text{Var}(\rmW_{i,+} \overline{\rvz}_{i-1}) + \frac{n_i}{2} \text{Var}(\rmW_{i,-} \underline{\rvz}_{i-1})\\
     &= \frac{n_i}{2} \bigg(\text{Var}(\rmW_{i,+}) \E(\overline{\rvz}_{i-1})^2\\
     & + \text{Var}(\overline{\rvz}_{i-1}) \E(\rmW_{i,+})^2 \!+\! \text{Var}(\rmW_{i,-}) \E(\underline{\rvz}_{i-1})^2 \!+\! \text{Var}(\underline{\rvz}_{i-1}) \E(\rmW_{i,-})^2\bigg)\\
 	&= \frac{\pi}{n_i}(1 - \frac{2}{\pi})\E(\overline{\rvz}_{i-1}^2)\!+\!\frac{2}{n_i}\Var(\overline{\rvz}_{i-1})
 	\!+\!\frac{\pi}{n_i}(1 - \frac{2}{\pi})\E(\underline{\rvz}_{i-1}^2)\!+\! \frac{2}{n_i}\Var(\underline{\rvz}_{i-1}).
 \end{align*}
 Note that $\E(\overline{\rvz}_{i}) \geq \E(\delta_{i})$ and we have made $\E(\delta_i)$ stable in each layer.
 Thus $\text{Var}(\overline{\rvh}_{i}) \geq \frac{n_i}{2}\text{Var}(\rmW_{i,+}) \E(\overline{\rvz}_{i-1})^2$ and will not vanish when the network goes deeper. 
 Also note that $n_i > 1$ in neural networks, and therefore $\text{Var}(\overline{\rvh}_{i})$ will not explode.
 The same analysis can also be applied to $\underline{\rvh}_{i}$. 
 
However, when we use the IBP initialization, variance of the standard forward value $\rvh_i$ will be smaller than that of Xavier and Kaiming Initialization. Following the analysis in \cite{He_2015_ICCV}, we have
\begin{align*}
 \Var{(\rvh_{i})} = \frac{n_i}{2}\Var{(\rmW_i)} \Var{(\rvh_{i-1})}.
\end{align*}
In IBP initialization, we have $\Var{(\rmW_i)} = \frac{2\pi}{n_i^2}$, and the variance of $\rvh_i$ can become smaller after going through each affine layer. 
Therefore, as mentioned in Section~\ref{sec:discussion}, simply adding IBP initialization may not finally improve the verified error, because it may harm the early warmup when $\epsilon$ is small and certified training is close to standard training.
In this paper, in addition to IBP initialization, we further add regularizers to stabilize certified bounds and the balance of ReLU neuron states, while the variance is stabilized by fully adding BN. 
The effect of these parts of our proposed method is discussed in Section~\ref{sec:discussion}.

\label{apd:proof_var}

\end{document}